\documentclass{article} 
\usepackage{iclr2025_conference,times}

\usepackage{amsmath,amsfonts,bm}

\def\eqref#1{equation~\ref{#1}}

\def\1{\bm{1}}

\DeclareMathAlphabet{\mathsfit}{\encodingdefault}{\sfdefault}{m}{sl}
\SetMathAlphabet{\mathsfit}{bold}{\encodingdefault}{\sfdefault}{bx}{n}

\usepackage{hyperref}
\usepackage{url}

\usepackage{times}
\usepackage{latexsym}
\usepackage{amsmath}
\usepackage{amssymb}
\usepackage{booktabs} 
\usepackage{enumitem}
\usepackage{graphicx}
\usepackage{pifont}
\usepackage{color}
\usepackage{titletoc}
\usepackage{colortbl}
\usepackage{wrapfig}  
\usepackage{lipsum} 
\usepackage{multirow}
\usepackage{listings}
\usepackage{array}
\usepackage{booktabs} 
\usepackage{caption}  
\usepackage{multicol}

\usepackage{inconsolata}
\usepackage{ulem}
\usepackage{array}
\usepackage{amsfonts}
\usepackage[ruled,linesnumbered,noend]{algorithm2e}
\usepackage{algpseudocode}
\usepackage{float}

\begin{document}

\newcommand{\ours}{MMIE}

\title{\includegraphics[height=0.5cm]{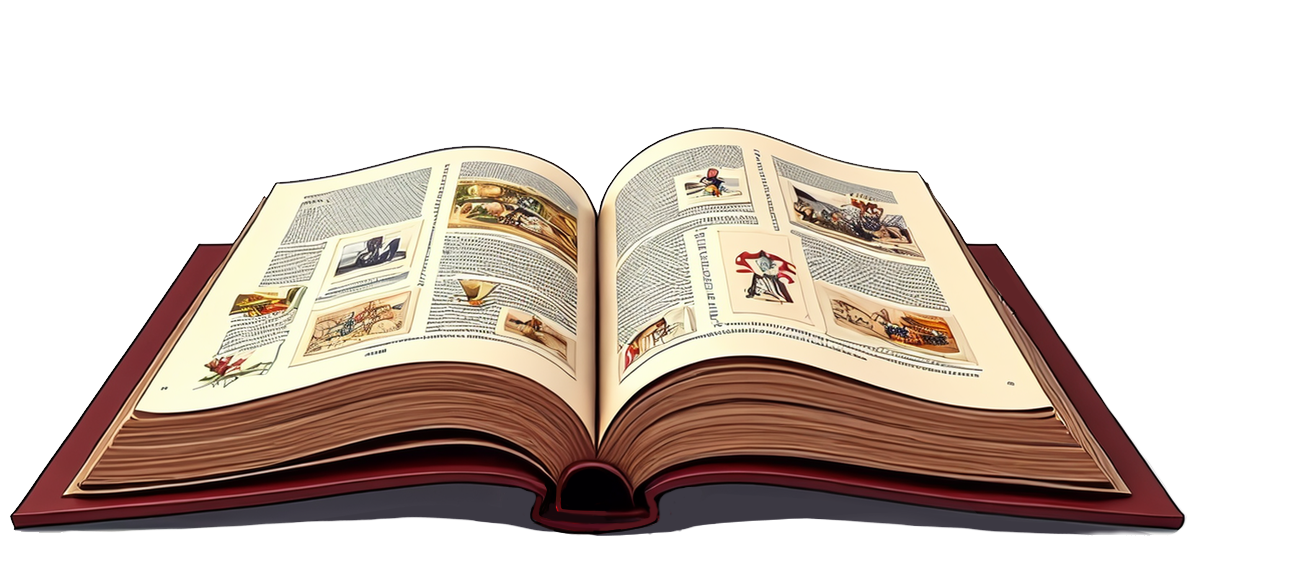} \hspace{-0.3cm} \ours: Massive Multimodal Interleaved Comprehension Benchmark for Large Vision-Language Models}

\iclrfinalcopy

\author{Peng Xia$^{1}$\thanks{Equal Contribution.}, Siwei Han$^{1*}$, Shi Qiu$^{1*}$, Yiyang Zhou$^{1}$, Zhaoyang Wang$^{1}$, Wenhao Zheng$^{1}$, \\ \textbf{Zhaorun Chen$^{3}$, Chenhang Cui$^{4}$, Mingyu Ding$^1$, Linjie Li$^2$, Lijuan Wang$^2$, Huaxiu Yao$^1$}\\ $^1$UNC-Chapel Hill, $^2$Microsoft Research, $^3$University of Chicago, $^4$NUS \\ \texttt{\{pxia,siweih,shiqiu,huaxiu\}@cs.unc.edu}}

\vspace{-2em}

\newcommand{\fix}{\marginpar{FIX}}
\newcommand{\new}{\marginpar{NEW}}

\maketitle
\vspace{-2em}
\begin{abstract}
Interleaved multimodal comprehension and generation, enabling models to produce and interpret both images and text in arbitrary sequences, have become a pivotal area in multimodal learning. Despite significant advancements, the evaluation of this capability remains insufficient. Existing benchmarks suffer from limitations in data scale, scope, and evaluation depth, while current evaluation metrics are often costly or biased, lacking in reliability for practical applications. To address these challenges, we introduce \ours, a large-scale knowledge-intensive benchmark for evaluating interleaved multimodal comprehension and generation in Large Vision-Language Models (LVLMs). \ours\ comprises 20K meticulously curated multimodal queries, spanning 3 categories, 12 fields, and 102 subfields, including mathematics, coding, physics, literature, health, and arts. It supports both interleaved inputs and outputs, offering a mix of multiple-choice and open-ended question formats to evaluate diverse competencies. Moreover, we propose a reliable automated evaluation metric, leveraging a scoring model fine-tuned with human-annotated data and systematic evaluation criteria, aimed at reducing bias and improving evaluation accuracy. Extensive experiments demonstrate the effectiveness of our benchmark and metrics in providing a comprehensive evaluation of interleaved LVLMs. Specifically, we evaluate eight LVLMs, revealing that even the best models show significant room for improvement, with most achieving only moderate results. We believe \ours\ will drive further advancements in the development of interleaved LVLMs. We publicly release our benchmark and code in \href{https://mmie-bench.github.io/}{https://mmie-bench.github.io/}. 
\end{abstract}
\begin{center}
    \textcolor{red}{Content warning: this paper contains content that may be inappropriate or offensive.}
\end{center}

\section{Introduction}
\textit{``True evaluation lies in the seamless interweaving of diverse modalities."}
\\

\vspace{-1em}
Multimodal learning has made remarkable progress with the development of Large Vision-Language Models (LVLMs)~\citep{liu2023visual,zhu2023minigpt,dai2023instructblip}, which are capable of handling diverse tasks that involve both images and text. Despite their advancements, most of these models are limited to multimodal tasks for text generation, such as visual question answering (VQA) and image captioning, which do not fully reflect the potential of multimodal capacity. To broaden their application, interleaved text-and-image generation has emerged as a critical area of research~\citep{liu2024holistic}. It requires models to generate images and text in any sequence, thereby enhancing the versatility and effectiveness of multimodal systems. It opens up possibilities for various complex applications, such as multi-step inference~\citep{lumathvista,kazemi2024remi}, multimodal situational analysis~\citep{yang2021visual}, and visual storytelling~\citep{huang2016visual}.

While recent LVLMs are evolving to support interleaved text-and-image generation~\citep{team2024chameleon,xie2024show,chern2024anole,zhou2024transfusion}, a comprehensive evaluation benchmark is still falling behind due to the following two challenges:
\begin{itemize}[leftmargin=*]
    \item \textbf{Difficulty in Constructing Modality-Coherent Benchmarks.} The first challenge lies in the difficulty of constructing modality-aligned multimodal datasets, where both the input and output contain images and text. Current benchmarks mainly focus on single-modality output tasks~\citep{fu2023mme,li2024seed,zhang2023magicbrush}, assessing only the quality of the generated image or text, without benchmarking the crucial connection between modalities, such as text-image coherence and consistency.
    Although a few datasets support the interleaved multimodal evaluation method for LVLMs~\citep{liu2024holistic}, their dataset is constrained by its limited scale and narrow query format, primarily focused on VQA tasks. 
    \item \textbf{Lack of Automated Evaluation Metric.} The second challenge is the lack of suitable automated evaluation metrics for interleaved generation. Human evaluation is costly and time-consuming, making it difficult to scale for practical applications. Current automated evaluation metrics typically assess either the quality of generated text (e.g., BLEU~\citep{papineni2002bleu}, BERTScore~\citep{zhangbertscore}) or the quality of generated images (e.g., FID~\citep{heusel2017gans}). While recent evaluation strategies, such as using CLIPScore~\citep{hessel2021clipscore}, and vision-language models (VLMs)~\citep{chen2023x,liu2024holistic}, can evaluate the connection between different modalities, they rely heavily on the pre-trained knowledge of specific models (e.g., CLIP training data) or follow rigid, human-defined rules. These approaches can introduce bias and uncertainty to some extent, often leading to inconsistent results~\citep{mahmoud2024sieve}.
\end{itemize}

\begin{figure}[t]
    \centering
    \includegraphics[width=0.9\linewidth]{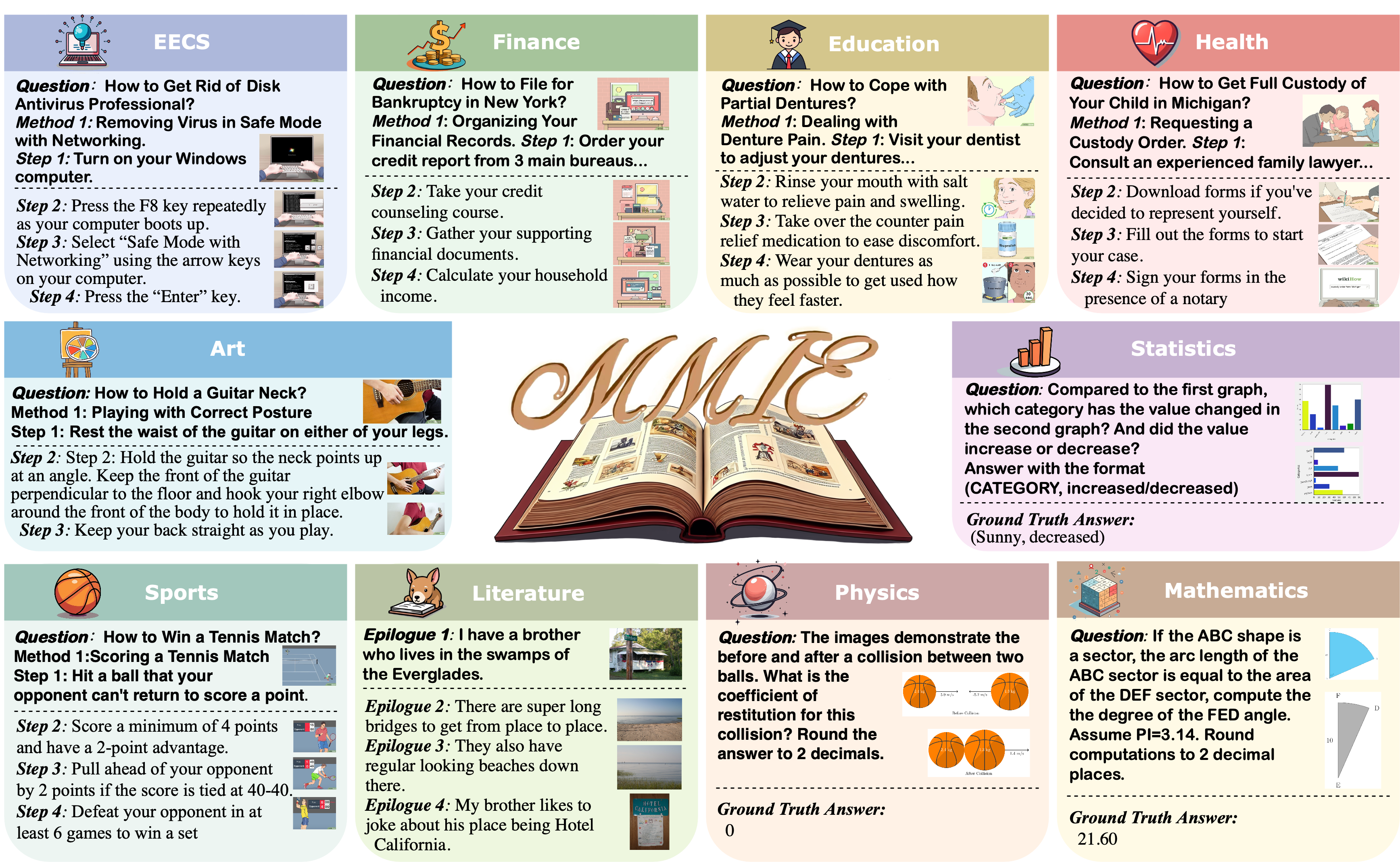}
    \caption{Typical samples from the \ours\ Benchmark showcase its support for multiple image inputs and outputs, with ground truth provided for every query. \ours\ evaluates models across diverse fields, ensuring a comprehensive evaluation of their capabilities.}
    \label{fig:overview}
    \vspace{-1em}
\end{figure}

To address these limitations, we introduce \textbf{\ours}, a \textbf{M}assive \textbf{M}ultimodal \textbf{I}nverleaved understanding \textbf{E}valuation benchmark for LVLMs with proposed reliable and automated metrics. \ours\ is curated from four multimodal datasets, involving 3 categories, 12 fields, and 102 subfields, including mathematics, physics, coding, statistics, literature, philosophy, education, finance, health, sports, art, and EECS (Electrical Engineering and Computer Science). The dataset comprises 20K multimodal questions, supporting both interleaved inputs and outputs. It features a mix of multiple-choice and open-ended question formats to evaluate a broad spectrum of competencies across various fields. As shown in Table \ref{tab:data_com}, \ours\ surpasses existing interleaved multimodal benchmark in both depth and width, particularly in addressing complex problem-solving and open-ended creative tasks. Based on the curated dataset, we further propose an automated metric powered by a scoring model. Specifically, we first design a comprehensive evaluation criteria for each category. Then, we curate a fine-grained, human-annotated scoring dataset and then use this dataset to fine-tune the InternVL-2~\citep{chen2024far} to obtain the scoring model. Using \ours, we evaluate four open-source interleaved multimodal LVLMs, as well as combinations of advanced LVLMs like GPT-4o with text-to-image generative models (e.g., Stable Diffusion 3~\citep{esser2024scaling}). Our key contributions are summarized as follows:

\begin{itemize}[leftmargin=*]
    \item We introduce the largest high-quality interleaved multimodal benchmark \ours\ for evaluating LVLMs, with the dataset to be publicly released.
    \item \ours\ presents significant challenges to LVLMs, with the best-performing model (e.g., GPT-4o + SDXL) achieving  a score of 65.47\%, highlighting substantial room for improvement.
    \item The proposed scoring model is reliable and has proven to be comparable to human evaluation.
\end{itemize}
\section{Related Work}
\textbf{Interleaved Multimodal Comprehension and Generation.} Multimodal learning has rapidly evolved, with substantial progress in integrating text and image modalities. Recent advancements in large vision-language models (LVLMs)~\citep{liu2023visual,zhu2023minigpt,zhu2024mmedpo,dai2023instructblip,xia2024mmed,xia2024rule}, either driven by the integration of diffusion models like Stable Diffusion~\citep{rombach2022high}, or using token-based mixed-modal structures like Chameleon~\citep{team2024chameleon} and Show-o~\citep{xie2024show}, have enabled models to not only understand and generate content across modalities, but also engage in interleaved multimodal comprehension and generation. As the demand for richer, more interactive AI grows, interleaved multimodal comprehension and generation is becoming an essential component in the development of next-generation LVLMs.

\textbf{LVLM Benchmarks.} Despite the rapid advancements in multimodal learning, evaluation benchmarks remain far from perfect. 
Previous benchmarks primarily focused on evaluating the base perception ability of LVLMs~\citep{lu2022learn,gurari2018vizwiz}, such as GQA~\citep{hudson2019gqa}, which lack the depth required to assess advanced reasoning. Recently, several high-quality evaluation benchmarks have been proposed to assess the reasoning ability of these models~\citep{li2024seed,zhang2023magicbrush,liu2023visual,liu2023mmbench,yu2023mm,xia2024cares,jiang2024mmsearch,zhang2024mathverse,zhang2024mathverse,zhang2024mavis,jiang2025mme}, such as MMMU~\citep{yue2024mmmu} and MME~\citep{fu2023mme}. However, these benchmark do not support interleaved image-and-text comprehension and generation. 
Large-scale interleaved multimodal datasets like MINT-1T~\citep{awadalla2024mint}, MANTIS~\citep{jiang2024mantis} and OBELICS~\citep{laurenccon2024obelics} have been developed primarily for pre-training models. However, they lack precise alignment between text and images, making them unsuitable for evaluation and benchmarking. A recent small-scale interleaved multimodal benchmark has been introduced~\citep{liu2024holistic}, but its limited data size and query quality hinder the comprehensiveness of its evaluation.
\ours\ fills this gap by offering a comprehensive evaluation framework that supports interleaved multimodal comprehension and generation. Our dataset includes a diverse set of queries among multiple domains. By evaluating both perceptual and generative capacity of LVLMs, it provides a more holistic assessment.

\begin{wraptable}{r}{0.45\textwidth}
\begin{center}
\scriptsize
\vspace{-2em}
\caption{Dataset statistics.}
\vspace{-1em}
\begin{tabular}{@{} l r r @{}}
\toprule
Statistic & Number & Percentage \\
\midrule
Total questions & 20103 & - \\
- Situational analysis & 5005 & 24.89\% \\
- Project-based learning & 11482 & 57.12\% \\
- Multi-step reasoning & 3616 & 17.99\% \\
Total Categories/Fields/Subfields & 3/12/102 & - \\
\midrule
Formats: & & \\
- Multiple-Choice Questions & 663 & 3.40\% \\
- Open-Ended Questions & 19340 & 96.60\% \\
\midrule
Questions with Images & 20103 & 100\% \\
Questions with answer label & 20103 & 100\% \\
Average question length & 76.0 & - \\
Average images per question & 1.32 & - \\
\bottomrule
\end{tabular}
\label{tab:sta}
\end{center}
\vspace{-1em}
\end{wraptable}

\textbf{Evaluation Metrics for Multimodal Tasks.} 
Traditional evaluation metrics, such as BLEU~\citep{papineni2002bleu}, BERTScore~\citep{zhangbertscore} for text quality, and FID~\citep{heusel2017gans} for image quality, are only suited to single-modality output tasks. 
Recent metrics, such as CLIPScore~\citep{hessel2021clipscore} and X-IQE~\citep{chen2023x}, have attempted to address this by introducing multimodal models to evaluate consistency between text and image. However, these metrics only measure alignment and fall short of offering a comprehensive assessment of output quality. Furthermore, many multimodal metrics depend on GPT-based models~\citep{liu2024holistic}, bringing uncontrollable bias to the whole evaluation system.
To overcome these drawbacks, we propose an automatic metric to minimises bias and provides a thorough analysis of the generated results.

\section{The \ours\ Benchmark}
\label{sec:bench}
\subsection{Overview}
\begin{wrapfigure}{r}{0.35\textwidth}
\vspace{-4em}
\begin{center}
    \includegraphics[width=0.35\textwidth]{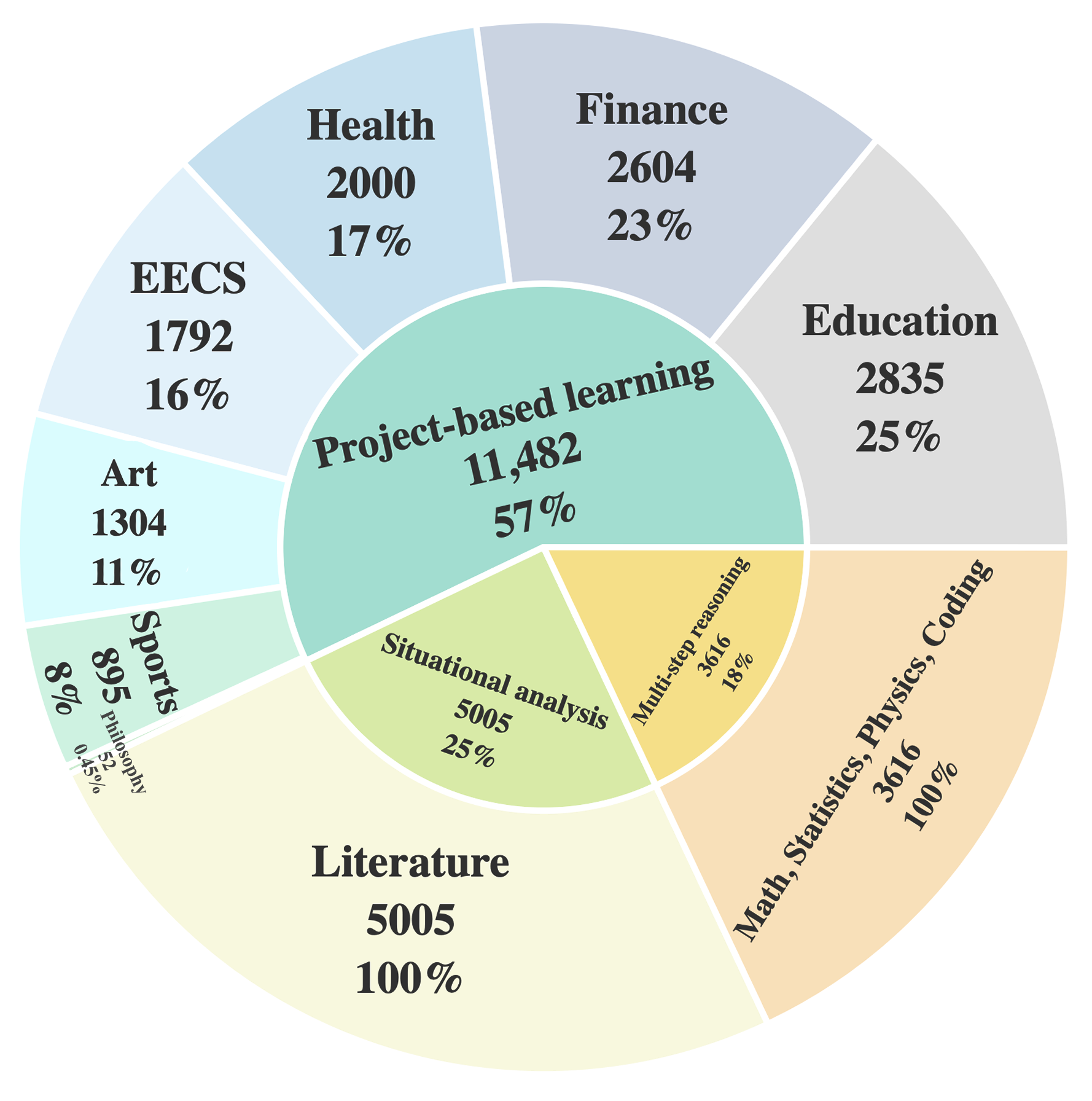}
\end{center}
\vspace{-1em}
\caption{Distribution of categories and fields in \ours.}
\label{fig:data_sta}
\vspace{-1em}
\end{wrapfigure}
In this section, we introduce \ours, a diverse and comprehensive benchmark for evaluating interleaved multimodal comprehension and generation across a broad scope of tasks. As shown in Table~\ref{tab:data_com}, \ours\ consists of 20,103 curated samples spanning 12 fields, including mathematics, physics, coding, statistics, literature, philosophy, education, finance, health, sports, art, and EECS. Each query is meticulously selected, filtered, and refined to ensure both high quality and relevance across the covered subjects. In addition, \ours\ emphasizes the evaluation of three essential competencies: perception, reasoning, and generation. Unlike previous benchmarks that evaluate the results from single modality~\citep{fu2023mme,yue2024mmmu,li2024mvbench} output, \ours\ is specifically designed to assess models’ capabilities in understanding and generating interleaved text and images in any sequence. This evaluation extends beyond basic perception by requiring models to engage in complex reasoning, leveraging subject-specific knowledge across different modalities.

\vspace{-1em}
\subsection{Dataset Curation}
\begin{wrapfigure}{r}{0.25\textwidth}
\vspace{-2em}
\begin{center}
    \includegraphics[width=0.99\linewidth]{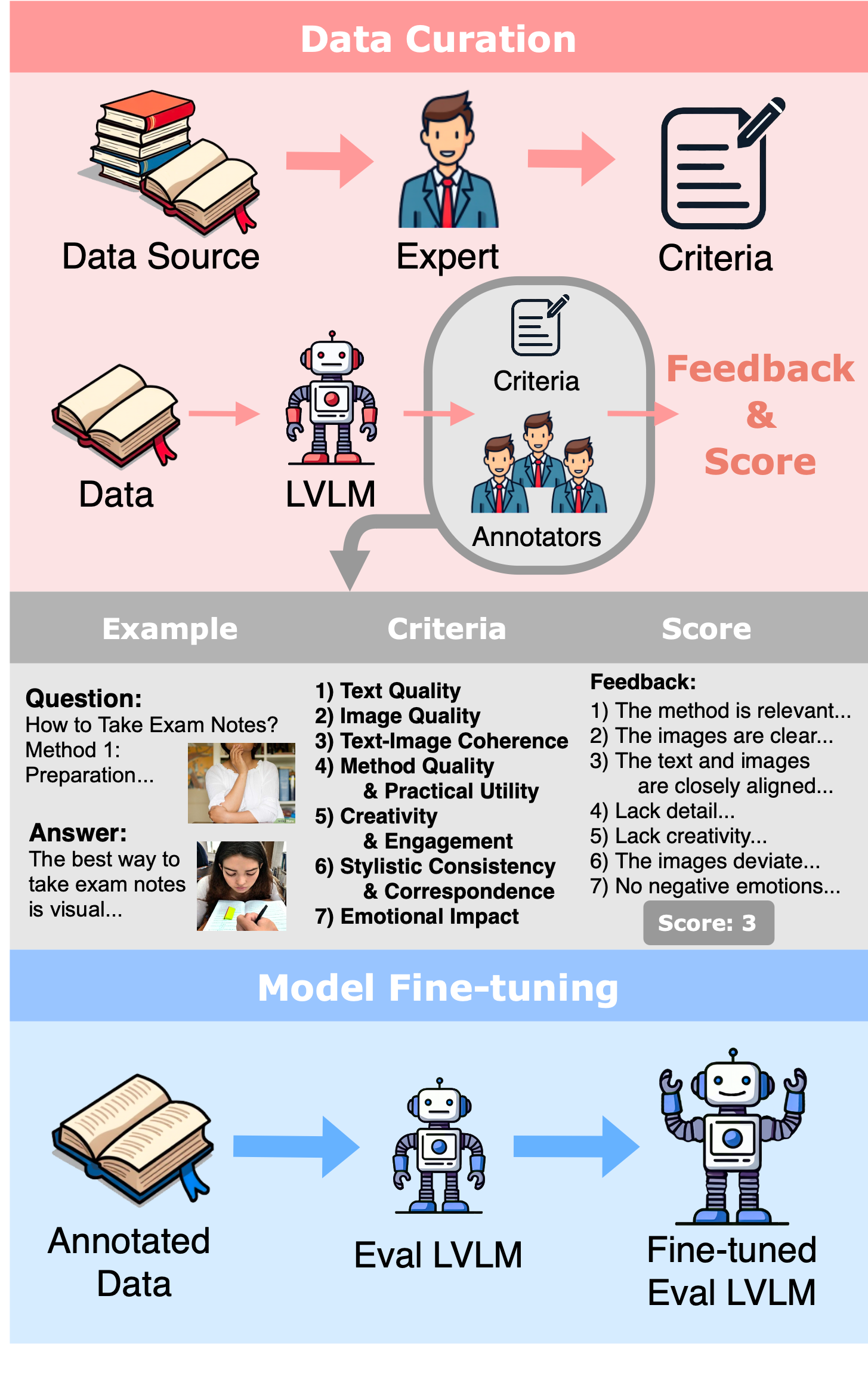}
\end{center}
\vspace{-1em}
\caption{Pipeline of the scoring model.}
\label{fig:score_model}
\vspace{-2em}
\end{wrapfigure}
The data curation process in \ours\ consists of two stages, each designed to ensure both comprehensive coverage and high-quality representation across various categories in our benchmark. We detail the process as follows:

In the first stage, we collect and restructure four multimodal datasets to align with the interleaved image-and-text format and categorize them into three categories -- situational analysis, project-based learning and multi-step reasoning, which are illustrated in Figure~\ref{fig:data_sta}. Specifically, for project-based learning, we extract data from Wikihow~\citep{yang2021visual}, which is originally designed for testing models' ability to choose the correct procedural steps based on given text and image contexts. We adapt it to the interleaved text-and-image format. For situational analysis, we draw samples from VIST~\citep{huang2016visual}, a naturally interleaved multimodal dataset designed for visual storytelling tasks, which challenges models to seamlessly integrate narrative text and images. Both situational analysis and project-based learning datasets feature interleaved inputs and outputs. To expand the benchmark with more complex and diverse tasks, we further introduce datasets focused on multi-step reasoning that support interleaved inputs. For this aspect, we source examples from MathVista~\citep{lumathvista} and ReMI~\citep{kazemi2024remi}, which together provide 3,600 questions covering topics from functions to statistics. The answer formats for these queries include multiple-choice questions (selecting one option from several choices) and open-ended questions (directly generating content).
After extracting samples from these four datasets, we merge and refine them into a cohesive benchmark by compacting, restructuring, and integrating questions from multiple sources, ensuring consistency with our evaluation objectives.

In the second stage, we implement a multi-step quality control process to ensure the integrity and consistency of the dataset. First, we apply lexical overlap and source URL similarity checks to identify and flag potential duplicate entries, which are then manually reviewed and removed. Next, each dataset is meticulously reviewed for formatting and typographical consistency to ensure adherence to a standardized structure. Discrepancies are corrected to maintain uniformity across the entire dataset. In total, we finally collect 20,103 instances across 12 fields, including mathematics, physics, coding, statistics, literature, philosophy, education, finance, health, sports, art, and EECS. Detailed categorization and dataset statistics are presented in Table~\ref{tab:sta}. For more information about dataset curation, please refer to Appendix~\ref{sec:data}.

\subsection{Automated Evaluation Metric}
\label{sec:metric}
As traditional metrics such as BLEU, BERTScore, and CLIP-Score fail to provide a thorough evaluation of the quality of multimodal outputs, existing benchmarks use the GPT-4 series as the scoring model, which may introduce inherent bias in the scoring process~\citep{liu2024holistic}. To ensure a comprehensive and unbiased evaluation of various LVLMs, as shown in Figure~\ref{fig:score_model}, we propose an automated evaluation metric powered by our fine-tuned LVLM to assist in scoring. Here, we choose InternVL-2-4B~\citep{chen2024far} as the foundation for our scoring system due to its strong performance in multimodal reasoning tasks and support for multi-image inputs. Furthermore, we fine-tune the InternVL-2-4B to mitigate potential bias.

Specifically, we first construct a high-quality multimodal scoring dataset that covers all aspects of our benchmark, accompanied by detailed scoring criteria and reference answers. In this process, we collect 800 responses from four LVLMs—MiniGPT-5~\citep{zheng2023minigpt}, EMU-2~\citep{sun2024generative}, GILL~\citep{koh2023generating}, and Anole~\citep{chern2024anole}. Based on the ground-truth labels, we define an evaluation standard using a six-point grading scale with clear criteria. A group of experts generates reference answers for each level and all score statistics are converted to percentage format. These criteria and reference answers together form a robust rubric for \ours. Following the rubric, human annotators rigorously score the responses. Detailed examples of the rubric and construction process are provided in Appendix~\ref{sec:cri} and Appendix~\ref{sec:data_score}. 

After constructing the scoring dataset, we fine-tune the InternVL-2-4B model and use the fine-tuned version as our scoring model. To validate its performance, we randomly select 200 new samples with human-scored labels and compare the results of our model with those of other scoring models. The results show that the fine-tuned model significantly improves alignment between human scores and our model-generated scores compared to other LVLMs, leading to more accurate and reliable evaluation across diverse tasks. We will discuss the experimental results in detail in Section~\ref{sec:exp_metric}.

\begin{table}[t]
    \centering
    \footnotesize
    \caption{The comparison between \ours\ with other LVLM benchmarks. Inter-I: interleaved input; Inter-O: interleaved output; Multi-I: multi-image for input; Multi-O: multi-image for output.}
    \setlength{\tabcolsep}{3pt}
    \vspace{-1em}
    \resizebox{\linewidth}{!}{
    \begin{tabular}{l|cccccccc}
    \toprule
       \textbf{Dataset} & \textbf{Data Scale} & \textbf{Inter-I} & \textbf{Inter-O} & \textbf{Multi-I} & \textbf{Multi-O} & \textbf{\#Num Domains} & \textbf{Answer Type} & \textbf{Metric} \\ \midrule
       HumanEval~\citep{chen2021evaluating} & 164 & No & No & No & No & 1 & Open & Pass$@$k \\
       GSM8K~\citep{cobbe2021training} & 8.5K & No & No & No & No & 1 & Open & Pass$@$k  \\
       MME~\citep{fu2023mme} & 2K & Yes & No & No & No & 4 & Multi-Choice & ACC \\
       MMBench~\citep{liu2023mmbench} & 3K & Yes & No & No & No & 6 & Multi-Choice & ACC  \\
       MM-Vet~\citep{yu2023mm} & 218 & Yes & No & No & No & 6 & Open & GPT-4 \\
       MagicBrush~\citep{zhang2023magicbrush} & 10K & Yes & No & No & No &  7 & Image Editing & CLIPScore \\ 
       MMMU~\citep{yue2024mmmu} & 11.5K & Yes & No & Yes & No & 30 & Multi-Choice & ACC \\
       MVBench~\citep{li2024mvbench} & 4K & Yes & No & Yes & No & 9 & Multi-Choice & ACC \\
       INTERLEAVEDBENCH~\citep{liu2024holistic} & 815 & Yes & Yes  & Yes & Yes &  10 & Open & GPT-4o\\ \midrule
       \textbf{\ours\ (Ours)} & \textbf{20K} & Yes & Yes  & Yes & Yes & 12 & Multi-Choice \& Open & Fine-tuned VLM \\ 
    \bottomrule
    \end{tabular}
    }
    \vspace{-1em}
    \label{tab:data_com}
\end{table}

\subsection{Comparison with Existing Multimodal Benchmarks}
\ours\ surpasses existing benchmarks in three key aspects. First, most previous multimodal benchmarks support only single-modality input or output, while \ours\ closes this gap by enabling interleaved text-and-image comprehension and generation. Our dataset ensures robust modality alignment, with multimodal question-answer pairs reconstructed into an interleaved text-and-image instruction format, followed by manual review to guarantee quality. Moreover, the scenarios reflect real-world applications, such as multimodal script generation, data chart analysis, and multimodal story generation. Second, compared to recent interleaved comprehension benchmarks~\citep{liu2024holistic}, \ours\ is larger in scale and covers a broader range of subjects, containing both reasoning and temporal understanding skills, allowing for a more comprehensive evaluation. Finally, \ours\ introduces a reliable scoring system powered by a fine-tuned LVLM, which significantly enhances the accuracy and reliability of scoring. Table~\ref{tab:data_com} highlights the differences between our benchmark and existing ones, demonstrating the advantages of \ours\ in terms of scale, diversity, and scoring methodology.
\section{Experiment}
\label{sec:exp}
\ours\ provides a systematic evaluation of existing open-source LVLMs supporting interleaved multimodal input and output (\textbf{interleaved LVLMs}), along with the integration of state-of-the-art LVLMs and text-to-image generative models (\textbf{integrated LVLMs}).
In this section, we aim to answer the following key questions: (1) Which interleaved LVLM performs best on \ours\ overall? (2) How effective are the integrated LVLMs? (3) Do the evaluated LVLMs show a preference for a certain field? and (4) How useful are our proposed model-powered metric compared with traditional metrics and other LVLM evaluation? 

\subsection{Experiment Setup}
\textbf{Baseline Models.} We first benchmark four open-source interleaved LVLMs. (1) MiniGPT-5~\citep{zheng2023minigpt}, a multimodal model combining MiniGPT-4 and Stable Diffusion, specialized for coherent image-text generation. (2) EMU-2~\citep{sun2024generative}, a 37B-parameter model excelling in in-context learning and multimodal reasoning, (3) GILL~\citep{koh2023generating}, a model specialized in generating and retrieving interleaved outputs, (4) Anole~\citep{chern2024anole}, based on Chameleon~\citep{team2024chameleon}, a model excelling in text quality, adds vision and multimodal generation capabilities.

To broaden the comparison, we also compare with integrated LVLMs consisting of text-output LVLMs (i.e., GPT-4o~\citep{achiam2023gpt}, Gemini-1.5~\citep{reid2024gemini}, LLaVA-v1.6-34b~\citep{liu2023visual} and Qwen-VL-2-72b~\citep{wang2024qwen2}) and text-to-image generative models (i.e., Openjourney~\citep{openjourney}, Stable Diffusion 3 Medium~\citep{esser2024scaling}, Stable Diffusion XL turbo, Flux.1-dev~\citep{flux.1-dev}). We provide the interleaved text-and-image input to the LVLM to generate text, and then feed this text to a text-to-image generative model to generate an image. The resulting multimodal output from this process is considered as interleaved output for evaluation. 


\noindent
\textbf{Human Annotators.} 
We organize a group of senior top-tier college students, contributing to the curation of the scoring dataset. To ensure thorough and consistent evaluations, we develop detailed criteria for each category of our benchmark (see Appendix~\ref{sec:cri} for details).

\noindent
\textbf{Evaluation Metrics.} We evaluate the performance of all models using our proposed metric in Section~\ref{sec:metric}, which is powered by our fine-tuned LVLM based on InternVL-2-4B~\citep{chen2024far}, to ensure reliable scoring.

\subsection{Main Results}
In this section, we present the comprehensive evaluation on our \ours\ benchmark. The detailed performance of interleaved LVLMs and integrated LVLMs is shown in Table~\ref{tab:open} and Table~\ref{tab:inte}, respectively. We summarize our key findings as follows:

\textbf{Challenging Evaluation and Promising Direction.} 
As illustrated in Table~\ref{tab:open}, all evaluated interleaved LVLMs show poor performance, with an average score of 50.80\%. Even when integrating advanced models such as GPT-4o and text-to-image generative models, as shown in  Table~\ref{tab:inte}, the best score (GPT-4o + SDXL) reached is 65.47\%. This highlights the high level of difficulty and the challenge posed by \ours. Interestingly, the latest interleaved LVLM Anole~\citep{chern2024anole} shows significant improvements over previous interleaved LVLMs, including MiniGPT-5, GILL and EMU-2, by 8.4\%, 7.0\%, 21.8\% in average score, respectively. This points to the growing potential of interleaved text-and-image models as a promising direction for future progress in multimodal comprehension and generation. To facilitate the broader adoption of \ours, we extract 1,000 samples to create a mini-set. Detailed results can be found in Appendix~\ref{sec:mini}.

\noindent
\textbf{Gap between Interleaved LVLMs and Integrated LVLMs.} Existing interleaved LVLMs are still quite limited. To enhance our evaluation and analysis on our benchmark, we integrate non-interleaved LVLMs with T2I models in our experiments. This integrated LVLMs approach significantly outperforms previous open-source interleaved LVLMs, improving performance by an average of 25.2\% across all categories. Specifically, the integrated models outperform the best performance of the interleaved model by 14.6\%, 26.3\%, and 16.1\% in situational analysis, project-based learning, and multi-step reasoning, respectively. Surprisingly, the integrated LVLMs perform exceptionally well in project-based learning, with all models based on LLaVA-34b achieving scores above 70\%. These findings suggest that combining the strong comprehension abilities of non-interleaved LVLMs with the generative power of T2I models offers a promising path for future research.

\begin{figure}[t]
    \centering
    \includegraphics[width=0.95\linewidth]{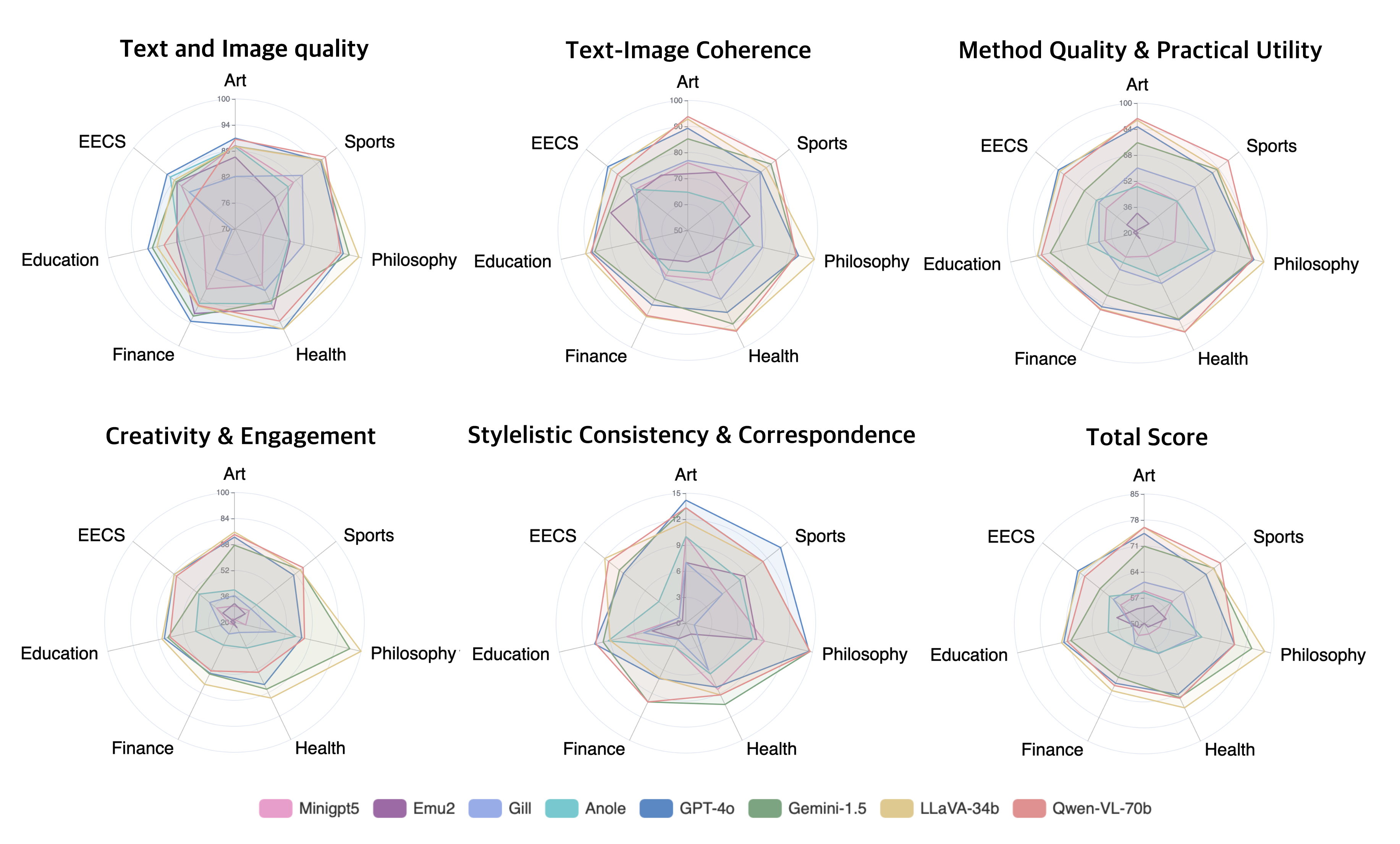}
    \vspace{-1em}
    \caption{The average and total scores of each model across the seven fields of project-based learning based on our criteria. We take the average of GPT-4o, Gemini-1.5, LLaVA-v1.6-34b and Qwen-VL-2-72b over the four text-to-image diffusion models.
    }
    \label{fig:main_results}
    \vspace{-2em}
\end{figure}

\noindent
\textbf{Model Performance across Different Fields.} 
As previously demonstrated in Table~\ref{tab:open} and Table~\ref{tab:inte}, model performance varies across different categories of data, achieving the best results in project-based learning and the lowest scores in situational analysis. This indicates that the model's performance differs depending on the category, likely due to inherent issues with the distribution of the training data.
For example, Anole~\citep{chern2024anole} scores 59.05\% in project-based learning data but only 48.95\% in situational analysis, suggesting it excels at creative, open-ended generation but falls short in handling detailed, discipline-specific knowledge. Delving into more fine-grained fields, as shown in Figure~\ref{fig:main_results}, different models exhibit preferences for certain fields of data. 
Among the seven fields of project-based learning, including education, finance, health, philosophy, sports, art and EECS, almost all models tend to perform well in areas that are easier to understand, such as philosophy, art and education, but face challenges in more complex fields requiring higher reasoning abilities, such as finance and EECS. Figure~\ref{fig:main_results} also shows a general gradual decline in scores for the criteria of text and image quality, text-image coherence, method quality and practical utility, creativity and engagement, stylistic consistency and correspondence, suggesting that there is a significant lack of text and image alignment and the ability to use interleaved output to solve real-world problems across all models. Detailed results can be found in Appendix~\ref{sec:fields}.

\begin{table}[t]
\centering
\footnotesize
\caption{Performance of the four open-source LVLMs supporting interleaved image-and-text input and output on \ours, shown as percentages.}
\setlength{\tabcolsep}{6pt}
\label{tab:open}
\vspace{-1em}
\resizebox{\linewidth}{!}{
\begin{tabular}{lccccc}
\toprule
\textbf{Model} & \textbf{Situational analysis} & \textbf{Project-based learning} & \textbf{Multi-step reasoning} & \textbf{AVG}\\
\midrule
MiniGPT-5~\citep{zheng2023minigpt}    & 47.63 & 55.12 & 42.17 & 50.92 \\
EMU-2~\citep{sun2024generative}                & 39.65 & 46.12 & 50.75 & 45.33 \\
GILL~\citep{koh2023generating} & 46.72 & 57.57 & 39.33 & 51.58 \\
Anole~\citep{chern2024anole}  & \textbf{48.95} & \textbf{59.05} & \textbf{51.72} & \textbf{55.22} \\
\bottomrule
\end{tabular}
}
\vspace{-1em}
\end{table}

\begin{table}[htbp]
\centering
\footnotesize
\caption{Comparison with state-of-the-art LVLMs integrated with text-to-image models, referred to as integrated LVLMs, evaluated on \ours. *: LLaVA only supports single-image input and all multi-image queries are thus skipped.}
\label{tab:inte}
\vspace{-1em}
\resizebox{\linewidth}{!}{
\begin{tabular}{clccccc}
\toprule
\textbf{LVLM} & \textbf{T2I Model} & \textbf{Situational analysis} & \textbf{Project-based learning} & \textbf{Multi-step reasoning} & \textbf{AVG} \\
\midrule
\multirow{4}{*}{GPT-4o} & Openjourney & 53.05 & 71.40 & \multirow{4}{*}{53.67} & 63.65 \\
& SD-3 & 53.00 & 71.20 & &63.52 & \\
& SD-XL & \textbf{56.12} & 73.25 & &\textbf{65.47} & \\
& Flux & 54.97 & 68.80 & &62.63 & \\
\midrule
\multirow{4}{*}{Gemini-1.5} & Openjourney & 48.08 & 67.93 & \multirow{4}{*}{\textbf{60.05}} & 61.57  \\
& SD-3 & 47.48 & 68.70 & & 61.87 & \\
& SD-XL & 49.43 & 71.85 & & 64.15 & \\
& Flux & 47.07 & 68.33 & & 61.55 & \\
\midrule
\multirow{4}{*}{LLaVA-34b} & Openjourney & 54.12 & 73.47 & \multirow{4}{*}{47.28*} & 63.93 \\
& SD-3 & 54.72 & 72.55 & & 63.57 & \\
& SD-XL & 55.97 & \textbf{74.60} & & 65.05 & \\
& Flux & 54.23 & 71.32 & & 62.73 & \\
\midrule
\multirow{4}{*}{Qwen2-VL-72b} & Openjourney & 52.73 & 71.63 & \multirow{4}{*}{55.63} & 64.05 \\
& SD-3 & 54.98 & 71.87 & & 64.75 & \\
& SD-XL & 52.58 & 73.57 & & 65.12 & \\
& Flux & 54.23 & 69.47 & & 63.18 & \\
\bottomrule
\end{tabular}
}
\end{table}

\subsection{How Consistent is Our Model-Powered Metric w.r.t Human Annotation?}
\label{sec:exp_metric}
In this section, we further validate the effectiveness of our proposed metric. Here, we conduct an experiment to evaluate its correlation with human annotations using several disparity and similarity metrics, i.e., cosine similarity, mean square error (MSE), mean absolute error (MAE), and Pearson coefficient. For comparison, we report results from traditional multimodal alignment metric (i.e., CLIPScore) and scores judged by LVLMs, including GPT-4o, which has already served as the metric in~\citep{liu2024holistic}. As shown in Table~\ref{tab:metric}, our metric demonstrates the closest alignment with human evaluation results significantly, proving to be the most reliable. Our scoring model effectively captures the multimodal features of both image and text sequences and judges them through complex reasoning precisely. In contrast, other LVLMs and CLIPScore tend to focus primarily on understanding the sequence information, but they fall short in grasping the relationships between the sequences and accurately judging the alignment between them. In summary, the experiments demonstrate that our metric is a robust and dependable standard for evaluating interleaved multimodal generation. We provide analyses of the scoring bias and generalization of the MMIE-Score in Appendices~\ref{sec:score_bias} and ~\ref{sec:score_general}, respectively.

\begin{table}[ht]
\centering
\footnotesize
\caption{Comparison of scoring LVLMs and traditional image-text alignment metric.}
\label{tab:metric}
\vspace{-1em}
\begin{tabular}{lccccc}
\toprule
\textbf{Models} & \textbf{Cosine Similarity} & \textbf{MSE} & \textbf{MAE} & \textbf{Pearson} \\
\midrule
Text-Image CLIPScore & 0.639 & 7.312 & 2.251 & 0.023 \\ \midrule
InternVL-2.0-4B   & 0.736 & 15.962 & 3.165 & 0.083 \\
Anole & 0.805 & 3.969 & 1.600 & 0.048 \\
GPT-4o                & 0.733 & 3.724 & 1.573 & 0.042 \\
\midrule
\textbf{Ours}     & \textbf{0.873} & \textbf{3.300}   & \textbf{1.444} & \textbf{0.113} \\
\bottomrule
\end{tabular}
\end{table}

\vspace{-2em}
\section{Error Analysis}
\begin{figure}[htbp]
    \centering
    \includegraphics[width=0.95\linewidth]{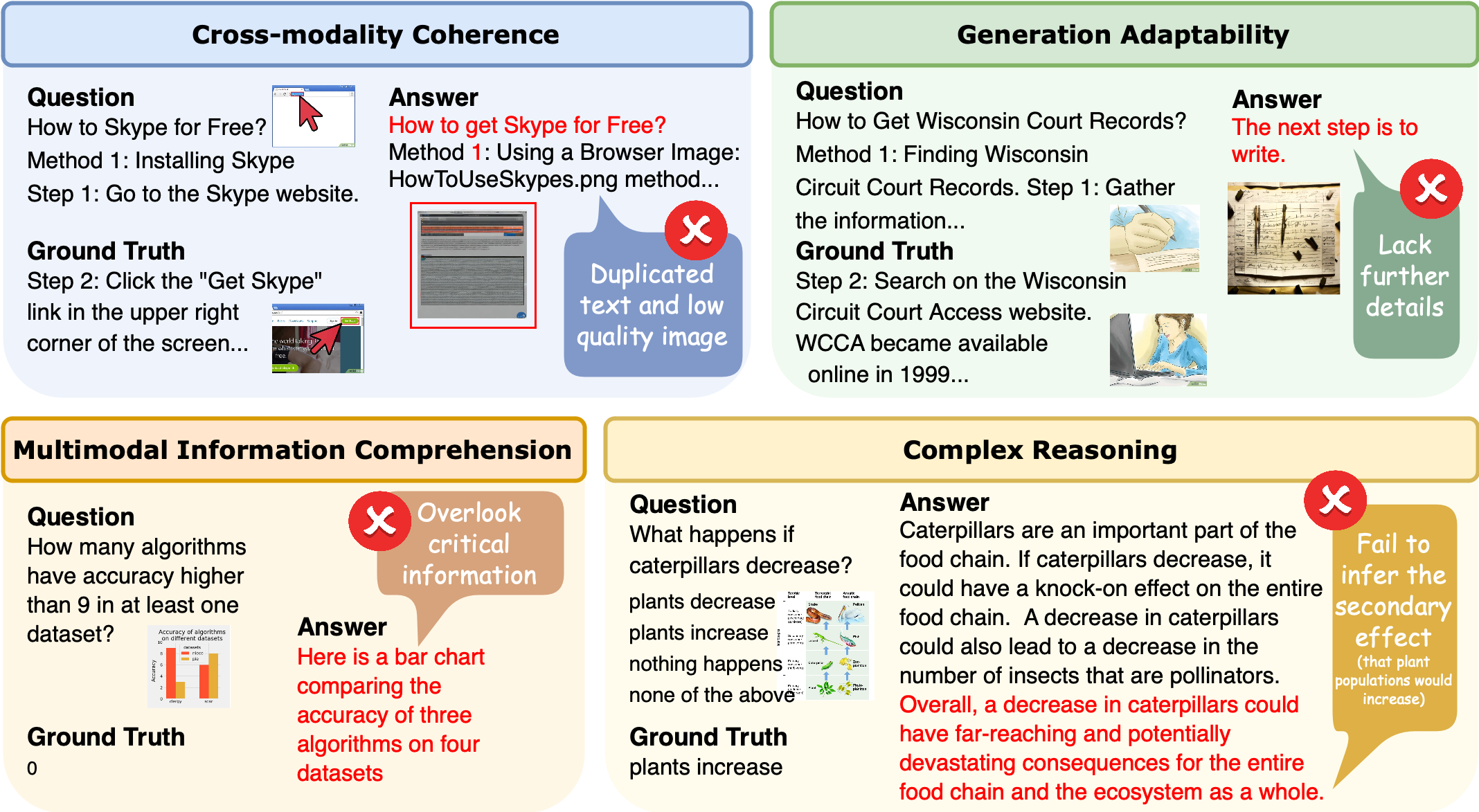}
    \caption{Examples of model failures. Four typical types of errors are introduced and categorized, namely incoherence between text and image generation, inflexibility in generated responses, poor comprehension of multimodal information, and inability to manage complex reasoning tasks.}
    \label{fig:fail}
    \vspace{-1em}
\end{figure}
This section offers a detailed analysis of the errors identified during the evaluation. We categorize the key challenges into two types: temporal understanding and reasoning ability. Specifically, temporal understanding issues refer to multimodal information comprehension and cross-modality coherence, while reasoning issues involve complex reasoning and generation capabilities. This analysis, drawn from expert annotators' observations during the scoring process, not only underscores the model's current limitations but also informs potential improvements for future development. Detailed examples can be found in Figure \ref{fig:fail}. More cases can be found in Appendix~\ref{sec:case}.

\subsection{Temporal Understanding Skill} The primary errors lie in \textit{cross-modality coherence and generation adaptability}. Many models struggle to generate images that accurately correspond to the accompanying text, resulting in severe information gaps, distortions, and redundancies. 

\noindent
\textbf{Cross-modality Coherence.} One of the most common errors is the incoherence between text and image generation. Due to deficiencies in multimodal alignment, the details in the generated images are often vague or entirely missing, making it difficult for them to align with the context described in the text. A typical example, as shown in Figure \ref{fig:fail}, involves the model understanding the ``Browser Image: HowToUseSkypes.png" method correctly and producing an accurate textual response. However, the corresponding image it generates consists of little more than blocks of color, lacking the necessary details to establish coherence and alignment with the text. 

\noindent
\textbf{Generation Adaptability.} Another significant error is the inflexibility of generated responses. For example, the model can only understand the given text and produce simple, detail-lacking responses. For example, in Figure \ref{fig:fail}, the model's reply merely contains the title ``the next step is to write" without further elaborating on the steps or process involved, which differs from the provided query example. This issue likely stems from a weakness in both text comprehension and generation.

\subsection{Reasoning Skill}
When evaluating the model's reasoning skills, the most prevalent error types are found in \textit{multimodal information comprehension and complex reasoning.} Notably, many models exhibit significant errors even in understanding interleaved information~\citep{jin2024impact,jin2024exploring,chen2024sharegpt4v,chen2024we,zhang2024map}, which inevitably leads to reasoning mistakes further down the process.

\noindent
\textbf{Multimodal Information Comprehension.} A key error in evaluating LVLMs' reasoning abilities is their difficulty in comprehending multimodal queries, especially in extracting visual information from images. A frequent issue arises when the model correctly interprets the textual components of a query but fails to fully understand the visual details in an image. For instance, in the case of a bar chart comparing four datasets by volume, where each dataset is represented by a bar with a corresponding height on the y-axis, the model might recognize the chart’s title and tags but overlook the critical information conveyed by the bars themselves—such as the relative sizes of the datasets. This highlights the model's tendency to focus on surface-level textual cues without delving into the deeper graphical meanings embedded in images. It also underscores a broader trend: LVLMs exhibit a strong bias toward processing text over extracting nuanced information from visual data and other non-textual modalities.

\noindent
\textbf{Complex Reasoning.} Another significant error is the model’s inability to handle complex reasoning tasks. As illustrated in Figure \ref{fig:fail}, the model demonstrates a pronounced weakness in multi-step inference. For example, in an impact analysis of a biological system, the model correctly predicts that a decrease in caterpillars would lead to a decline in bird populations but fails to infer the secondary effect—that plant populations would increase. Another instance is seen in arithmetic problems, where the model makes clear mistakes, such as failing to calculate the exact length of a triangle. These examples underscore the need to strengthen the model's capacity for multi-step reasoning, making it more robust and reliable in handling complex tasks.

\section{Conclusion}
This paper introduces \ours, a large-scale, diverse benchmark for interleaved image-and-text understanding and generation. Spanning a wide range of fields, \ours\ provides a comprehensive evaluation framework for interleaved multimodal understanding and generation, featuring 20K queries. The dataset, which covers a wide range of fields, ensures high-quality evaluation of LVLMs across various dimensions. Furthermore, our proposed model-powered metric effectively evaluates the quality of output image-text information based on the input image-text context. Our extensive experiments further demonstrate that the metrics we propose provide robust, human-like evaluation performance, significantly reducing errors and biases. Despite this, we observe that existing models underperform, particularly in complex and deeply interleaved multimodal tasks, highlighting the challenges and opportunities that lie ahead in this domain.
\section*{Ethics Statement}
This paper focuses on the evaluation of interleaved large vision-language models. A newly constructed human-annotated dataset was used to fine-tune the scoring model.  The dataset was curated following ethical guidelines to ensure that no sensitive information is included and to minimize bias during the annotation process. The evaluation process aims to be transparent and reproducible, adhering to high standards of research integrity and ethical conduct. No personally identifiable data was collected or processed.

\section*{Reproducibility Statement}
To ensure the reproducibility of our results, we have made considerable efforts to provide all necessary details and materials. Specifically, we have included a comprehensive description of the dataset creation process in Section~\ref{sec:bench}, including annotation guidelines and data collection methods, and further elaborated in Appendix~\ref{sec:data}. The benchmark and evaluation procedures are described in detail in Section~\ref{sec:exp}, with the metrics used clearly defined to facilitate independent verification.

\section*{Acknowledgement}
We would like to thank Danni Luo, Rundong Liu, Anqi Lv, Xin Xie, Bocheng Feng, Huanyan Ye, Shi'en Wan, Kaiyao Chu, Danning Li, Yiran Li, Siyuan Chen to participate in refining the criteria and Kaiyuan Zheng for his assistance in model selection. This work is partially supported by Cisco Faculty Research Award.

\bibliography{main}
\bibliographystyle{iclr2025_conference}

\appendix
\section{Appendix}
\subsection{Related Datasets and Metrics}
\label{sec:data}
\begin{itemize}
    \item \textbf{VIST}~\citep{huang2016visual} is a high-quality multimodal dataset for visual storytelling and interleaved text-and-image generation. It contains 5K individual stories containing both image and text in arbitrary orders.
    \item \textbf{ReMI}~\citep{kazemi2024remi} is a dataset designed to evaluate large language models (LLMs) on multi-image reasoning across diverse tasks like math, physics, logic, and spatial reasoning. It highlights key challenges in reasoning with multiple images, revealing a significant gap between current LLM performance and human proficiency.
    \item \textbf{MathVista}~\citep{lumathvista} is a benchmark designed to assess mathematical reasoning in visual contexts. MathVista comprises 6,141 examples from 28 existing multimodal datasets and three new datasets (IQTest, FunctionQA, and PaperQA).
    \item \textbf{Wikihow-VGSI}~\citep{yang2021visual} is a benchmark designed for multimodal comprehension, featuring a diverse array of examples sourced from WikiHow, primarily centered on methods to achieve specific goals. Initially released as a choice dataset, it includes multiple images and text presented in a selected order within each example, enhancing its potential for practical applications.
    \item \textbf{CLIPScore}~\citep{hessel2021clipscore} is a reference-free metric for evaluating image captioning by leveraging CLIP, a cross-modal model trained on 400M image-caption pairs. While effective for literal descriptions and tasks like alt-text rating, CLIPScore is less suited for news captions requiring deep contextual knowledge.
\end{itemize}

\subsection{Overview of Baseline Models}
\begin{itemize}
    \item \textbf{MiniGPT-5}~\citep{zheng2023minigpt} combines pretrained multimodal large language model MiniGPT-4 and image-generation model Stable Diffusion to implement multimodal inputs and outputs. It employs unique visual tokens called "generative vokens" that connect the textual and visual domains throughout the training process.
    \item \textbf{EMU-2}~\citep{sun2024generative} is a 37B generative multimodal model. The base model is then fine-tuned with converSituational analysistional data and image data separately to yield multimodal language model Emu2-Chat and visual generation model Emu2-Gen. In our experiment, we use a pipeline of Emu2-Chat and Emu2-Gen.
    \item \textbf{GILL}~\citep{koh2023generating} uses a mapping network to translate hidden representations of text into the embedding space of the visual models. It combines text-only LLMs with pre-trained image encoder and decoder models to process arbitrarily mixed image and text inputs and generate text combined with image embedding.
    \item \textbf{Anole}~\citep{chern2024anole} is a model fine-tuned on Meta Chameleon, relying solely on transformers. It facilitated Chameleon's image generation and multimodal generation capabilities by fine-tuning only the logits corresponding to image token ids in transformer’s output head layer.
    \item \textbf{GPT-4o}~\citep{achiam2023gpt} is an advanced language model developed by OpenAI, designed to enhance the capabilities of the GPT-4 architecture. It integrates innovations in transformer models and multi-modal processing, making it capable of handling both text and visual inputs.
    \item \textbf{Gemini-1.5}~\citep{reid2024gemini} is a large language model developed by Google AI, trained on a massive dataset of text and code. It can process and analyze both text and images input.
    \item \textbf{LLaVA-34b}~\citep{liu2023visual} is an end-to-end trained model that connects a vision encoder and an LLM for general-purpose visual and language understanding. The 34b version uses Hermes-Yi-34B as its LLM. However, it doesn't support multiple images as input, making it unable to cope with some of the expert level tasks in our \ours.
    \item \textbf{Qwen2-VL-72b}~\citep{wang2024qwen2}, is a multimodal version of the large model series Qwen, proposed by Alibaba Cloud. It is designed to process and understand multiple types of input, including text, images, and audio.
    \item \textbf{Openjourney}~\citep{openjourney} is a generative AI model designed specifically for creating high-quality images based on text prompts. It is a variant of the Stable Diffusion model, optimized for artistic and creative visual generation fine-tuned on Midjourney images.
    \item \textbf{Stable Diffusion 3 Medium}~\citep{esser2024scaling} is a text-to-image model developed by Stability AI. It's a powerful tool that can generate high-quality images from simple text descriptions, which produces images with greater detail, clarity, and overall quality.
    \item \textbf{Stable Diffusion XL turbo}~\citep{esser2024scaling} is an enhanced version of the Stable Diffusion XL model, optimized for faster image generation without compromising quality. Designed for efficiency, it allows users to create highly detailed and vivid images from text prompts at a significantly accelerated pace.
    \item \textbf{Flux.1-dev}~\citep{flux.1-dev} is a text-to-image and image-to-image model developed by Black Forest Labs. It is a 12 billion parameter rectified flow transformer capable of generating images from text descriptions. FLUX.1-dev is a guidance-distilled variant of the base FLUX.1 model, and is designed to be more efficient and easier to use.
\end{itemize}

\subsection{Data Curation for Fine-Tuning and Evaluating MMIE-Score}
\label{sec:data_score}
\begin{figure}[htbp]
    \centering
    \includegraphics[width=0.93\linewidth]{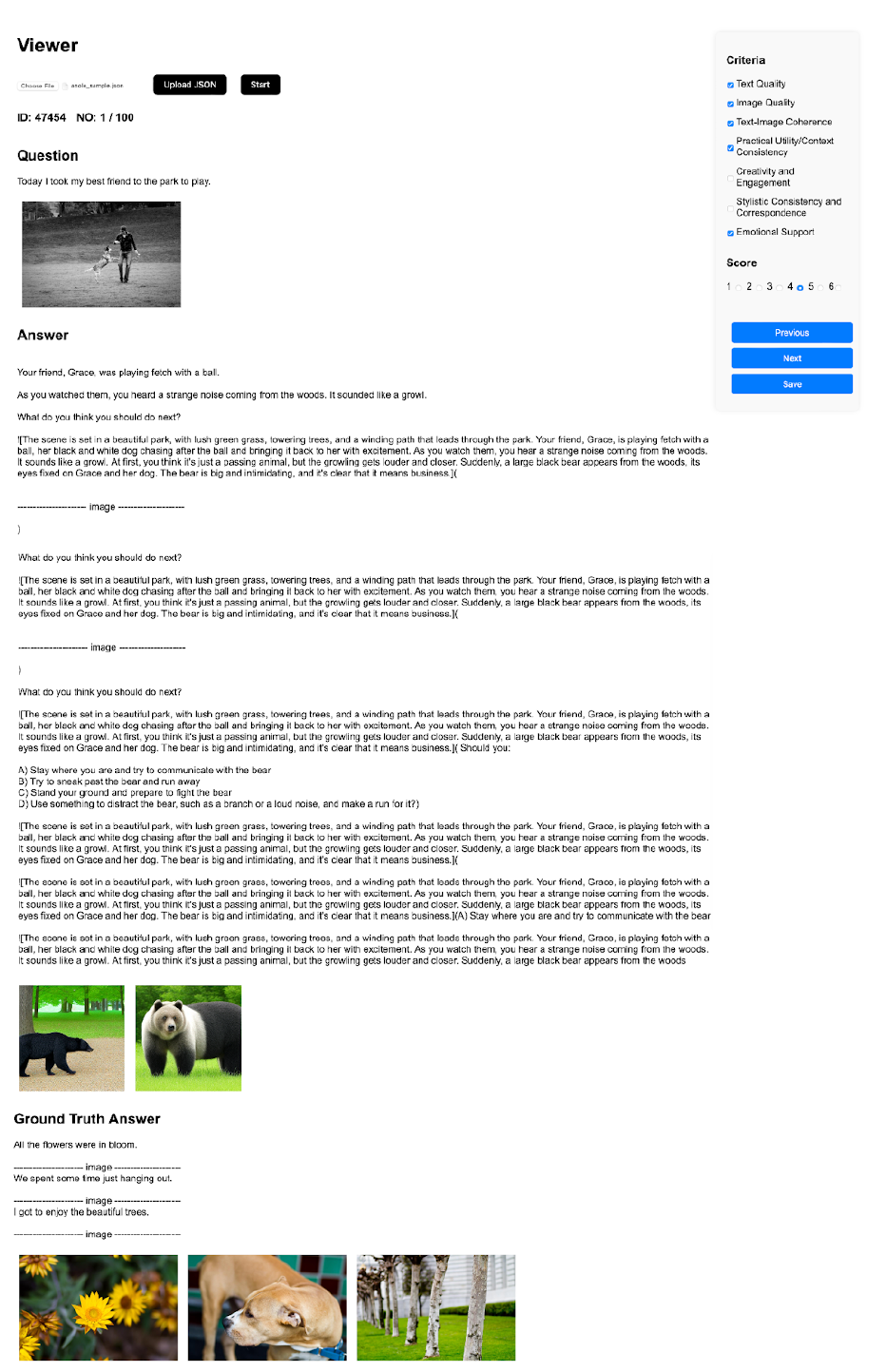}
    \caption{Screenshots of the graphical annotation tool. In this tool, the samples are displayed on the left, while a floating panel on the right allows annotators to score the samples. Annotators can check different criteria, and the final cumulative score is calculated accordingly. Checking a criterion indicates that the sample meets that specific criterion.}
    \label{fig:tool}
    \vspace{-1em}
\end{figure}

We randomly select examples from each field for human annotation, including the original inputs (images and questions), ground truth, and responses from evaluation models. For each category, we develop comprehensive and detailed criteria with scoring standards. We first annotate 20 examples, providing specific examples for each score as references. To facilitate the annotation process, we design a graphical annotation tool (See Figure~\ref{fig:tool}). Finally, we create a dataset of 1K examples with evaluation scores through human annotation, with 800 examples used for fine-tuning the scoring model and 200 examples for evaluating the scoring model. 

\subsection{Analysis of Bias in Scoring Model}
\label{sec:score_bias}
We present several comparison results of our scoring model with other baselines across different categories and model types (Interleaved and Integrated LVLMs). As shown in Table~\ref{tab:bias}, the results show that although our model exhibits slightly varying performance across different categories, for example Cosine Similarity scores are higher for Situational Analysis (SA) and Multi-Step Reasoning (MSR) category and Pearson scores are higher for Project-Based Learning (PBL) category, the biases remain little. Overall, our MMIE-Score consistently outperforms other baselines.

\begin{table*}[htbp]
\centering
\footnotesize
\caption{Comparison of scoring LVLMs and traditional image-text alignment metrics across different models and categories.}
\label{tab:bias}
\resizebox{\linewidth}{!}{
\begin{tabular}{llcccc}
\toprule
\textbf{Category} & \textbf{Models} & \textbf{Cosine Similarity $\uparrow$} & \textbf{MSE $\downarrow$} & \textbf{MAE $\downarrow$} & \textbf{Pearson $\uparrow$} \\
\midrule
\multirow{5}{*}{Situational Analysis (Interleaved)} 
& Text-Image CLIPScore & 0.604 & 6.710 & 2.057 & 0.022 \\
& InternVL-2.0-4B & 0.691 & 14.001 & 3.382 & 0.094 \\
& Anole & 0.867 & 3.973 & 1.579 & 0.045 \\
& GPT-4o & 0.718 & 4.195 & 1.573 & 0.042 \\
& \textbf{MMIE-Score (Ours)} & \textbf{0.895} & \textbf{3.547} & \textbf{1.502} & \textbf{0.098} \\
\midrule
\multirow{5}{*}{Project-based Learning (Interleaved)} 
& Text-Image CLIPScore & 0.612 & 7.669 & 2.197 & 0.022 \\
& InternVL-2.0-4B & 0.654 & 16.560 & 3.499 & 0.072 \\
& Anole & 0.689 & 4.423 & 1.566 & 0.047 \\
& GPT-4o & 0.661 & 3.837 & 1.670 & 0.045 \\
& \textbf{MMIE-Score (Ours)} & \textbf{0.760} & \textbf{3.163} & \textbf{1.496} & \textbf{0.114} \\
\midrule
\multirow{5}{*}{Multi-step Reasoning (Interleaved)} 
& Text-Image CLIPScore & - & - & - & - \\
& InternVL-2.0-4B & 0.770 & 16.050 & 3.393 & 0.084 \\
& Anole & 0.739 & 3.612 & 1.615 & 0.054 \\
& GPT-4o & 0.798 & 3.985 & 1.674 & 0.045 \\
& \textbf{MMIE-Score (Ours)} & \textbf{0.814} & \textbf{3.767} & \textbf{1.347} & \textbf{0.106} \\
\midrule
\multirow{5}{*}{Situational Analysis (Integrated)} 
& Text-Image CLIPScore & 0.640 & 7.701 & 2.184 & 0.023 \\
& InternVL-2.0-4B & 0.695 & 13.960 & 3.432 & 0.073 \\
& Anole & 0.823 & 4.222 & 1.408 & 0.051 \\
& GPT-4o & 0.763 & 3.707 & 1.521 & 0.039 \\
& \textbf{MMIE-Score (Ours)} & \textbf{0.843} & \textbf{2.811} & \textbf{1.384} & \textbf{0.093} \\
\midrule
\multirow{5}{*}{Project-based Learning (Integrated)} 
& Text-Image CLIPScore & 0.551 & 7.214 & 1.990 & 0.023 \\
& InternVL-2.0-4B & 0.759 & 15.250 & 3.075 & 0.094 \\
& Anole & 0.691 & 4.295 & 1.718 & 0.050 \\
& GPT-4o & 0.748 & 3.958 & 1.782 & 0.044 \\
& \textbf{MMIE-Score (Ours)} & \textbf{0.795} & \textbf{3.388} & \textbf{1.432} & \textbf{0.122} \\
\midrule
\multirow{5}{*}{Multi-step Reasoning (Integrated)} 
& Text-Image CLIPScore & - & - & - & - \\
& InternVL-2.0-4B & 0.688 & 17.451 & 3.590 & 0.092 \\
& Anole & 0.713 & 3.713 & 1.727 & 0.054 \\
& GPT-4o & 0.753 & 3.594 & 1.491 & 0.036 \\
& \textbf{MMIE-Score (Ours)} & \textbf{0.825} & \textbf{3.112} & \textbf{1.583} & \textbf{0.097} \\
\bottomrule
\end{tabular}
}
\end{table*}

\subsection{Mini Subset for Consistent yet Quicker Evaluation}
\label{sec:mini}
The fully 20k data size surely ensures the comprehensiveness of our dataset, but it does come with time costs. We resample evenly from each category and field, and construct a subset of 1000 samples. We re-ran the models on this subset and used MMIE-Score for scoring. As shown in Table~\ref{tab:subset1} and Table~\ref{tab:subset2}, the performance of the models on our subset is consistent with the results from the full dataset.

\begin{table}[htbp]
\centering
\footnotesize
\caption{Performance of the four open-source LVLMs supporting interleaved image-and-text input and output on \ours's 1K subset.}
\setlength{\tabcolsep}{6pt}
\label{tab:subset1}
\vspace{-1em}
\resizebox{\linewidth}{!}{
\begin{tabular}{lcccc}
\toprule
\textbf{Model} & \textbf{Situational analysis} & \textbf{Project-based learning} & \textbf{Multi-step reasoning} & \textbf{AVG} \\
\midrule
MiniGPT-5 & 46.26 & 56.53 & 46.06 & 52.09 \\
EMU-2     & 34.44 & 52.81 & 48.91 & 47.54 \\
GILL      & 48.48 & 59.49 & 35.88 & 52.50 \\
Anole     & \textbf{50.26} & \textbf{60.70} & \textbf{50.11} & \textbf{56.20} \\
\bottomrule
\end{tabular}
}
\vspace{-1em}
\end{table}

\begin{table}[htbp]
\centering
\footnotesize
\caption{Comparison with integrated LVLMs, evaluated on \ours's 1K subset. *: LLaVA only supports single-image input and all multi-image queries are thus skipped.}
\label{tab:subset2}
\vspace{-1em}
\resizebox{\linewidth}{!}{
\begin{tabular}{clccccc}
\toprule
\textbf{LVLM} & \textbf{T2I Model} & \textbf{Situational analysis} & \textbf{Project-based learning} & \textbf{Multi-step reasoning} & \textbf{AVG} \\
\midrule
\multirow{4}{*}{GPT-4o} & Openjourney & 56.00 & 67.81 & \multirow{4}{*}{56.70} & 62.87 \\
& SD-3 & 51.04 & 69.48 &  & 62.59 \\
& SD-XL & 55.07 & 73.92 &  & 66.13 \\
& Flux & \textbf{57.25} & 70.69 &  & 64.83 \\
\midrule
\multirow{4}{*}{Gemini-1.5} & Openjourney & 48.25 & 71.16 & \multirow{4}{*}{\textbf{60.98}} & 63.63 \\
& SD-3 & 45.88 & 70.12 &  & 62.44 \\
& SD-XL & 47.53 & 73.40 &  & 64.73 \\
& Flux & 47.23 & 70.05 &  & 62.74 \\
\midrule
\multirow{4}{*}{LLaVA-34b} & Openjourney & 54.84 & 72.87 & \multirow{4}{*}{47.58*} & 63.83 \\
& SD-3 & 56.35 & 72.71 &  & 64.12 \\
& SD-XL & 53.52 & \textbf{77.79} &  & 66.31 \\
& Flux & 55.17 & 68.27 &  & 61.29 \\
\midrule
\multirow{4}{*}{Qwen2-VL-72b} & Openjourney & 54.46 & 72.49 & \multirow{4}{*}{56.69} & 65.16 \\
& SD-3 & 52.25 & 74.98 &  & 66.03 \\
& SD-XL & 54.45 & 75.03 &  & \textbf{66.61} \\
& Flux & 55.76 & 67.19 &  & 62.46 \\
\bottomrule
\end{tabular}
}
\end{table}

\subsection{Visual Component Importance}
Our dataset curation and filtering process ensures that all images included in the examples contribute meaningfully to the overall task. For instance, in tasks like visual storytelling, even when images serve only an illustrative purpose, they still impact the overall output quality. We conduct a comparative experiment on 100 samples from \ours\ dataset to evaluate the difference in performance between interleaved generation (text and images) and text-only generation for the same input and same evaluated LVLM (GPT-4o + SDXL). The evaluation is scored using MMIE-Score and GPT-4o. As shown in Table~\ref{tab:image_generation_comparison}, results show that when the model outputs included both text and images, the overall quality is superior to text-only outputs. This ensures that the inclusion of images follows reasonable and well-defined criteria.

\begin{table}[htbp]
\centering
\footnotesize
\caption{Comparison of GPT-4o + SDXL's average score with and without image generation, evaluated by GPT-4o and
MMIE-Score.}
\label{tab:image_generation_comparison}
\begin{tabular}{lcc}
\toprule
\textbf{Model}     & \textbf{w/o image generation} & \textbf{w/ image generation} \\ \midrule
GPT-4o             & 60.90                         & \textbf{71.24}              \\ 
MMIE-Score (\textbf{Ours})         & 53.46                         & \textbf{65.47}              \\ \bottomrule
\end{tabular}
\end{table}

\subsection{Performance across Fields}
\label{sec:fields}
We provide detailed results in Table~\ref{tab:field} for the model's performance across the following 12 fields. In the Project-based learning (PBL) category, among the seven fields, most models exhibit slightly lower performance in finance and education, while achieving better results in art, sports, and philosophy. In the Multi-step reasoning (MSR) category, across the four fields, models generally perform worse in coding and statistics but demonstrate stronger performance in mathematics and physics.

\begin{table}[htbp]
\centering
\footnotesize
\caption{Performance comparison across various fields for different LVLMs. Since the Multi-step reasoning category (i.e., mathematics, physics, coding, statistics) do not require image outputs, the scores of integrated LVLMs for this category remain consistent.}
\label{tab:field}
\vspace{-1em}
\resizebox{\linewidth}{!}{
\begin{tabular}{lcccccccccccc}
\toprule
\textbf{Model} & \textbf{Art} & \textbf{EECS} & \textbf{Education} & \textbf{Finance} & \textbf{Health} & \textbf{Philosophy} & \textbf{Sports} & \textbf{Literature} & \textbf{Mathematics} & \textbf{Physics} & \textbf{Coding} & \textbf{Statistics} \\
\midrule
MiniGPT-5 & 58.90 & 58.13 & 53.17 & 53.47 & 53.00 & 54.18 & 59.80 & 47.63 & 46.53 & 42.75 & 40.24 & 39.17 \\
Emu-2 & 50.23 & 48.65 & 43.90 & 44.55 & 45.52 & 42.47 & 48.35 & 39.65 & 54.93 & 51.10 & 47.66 & 49.31 \\
Gill & 61.27 & 60.60 & 53.12 & 55.80 & 58.92 & 64.53 & 63.67 & 46.72 & 38.11 & 42.73 & 38.05 & 38.43 \\
Anole & 58.37 & 61.95 & 59.93 & 56.62 & 58.73 & 65.95 & 59.10 & 48.95 & 47.37 & 53.92 & 49.81 & 55.79 \\
GPT+OpenJourney & 74.30 & 74.12 & 70.92 & 69.77 & 70.07 & 79.17 & 71.98 & 53.05 & 58.07 & 51.63 & 52.60 & 52.37  \\
GPT+SD3 & 74.02 & 74.97 & 70.72 & 66.23 & 72.88 & 75.00 & 72.93 & 53.00 & 58.07 & 51.63 & 52.60 & 52.37 \\
GPT+SDXL & 74.87 & 72.57 & 75.05 & 71.87 & 73.75 & 66.67 & 70.65 & 56.12 &58.07 & 51.63 & 52.60 & 52.37 \\
GPT+Flux & 74.37 & 69.83 & 72.10 & 63.15 & 67.63 & 79.17 & 69.88 & 54.97 &58.07 & 51.63 & 52.60 & 52.37 \\
Gemini+OpenJourney & 68.47 & 65.28 & 67.48 & 67.35 & 68.67 & 76.18 & 73.65 & 48.08 & 61.83 & 65.02 & 55.85 & 57.49  \\
Gemini+SD3 & 69.78 & 67.17 & 69.57 & 64.02 & 71.87 & 83.33 & 72.55 & 47.48 & 61.83 & 65.02 & 55.85 & 57.49 \\
Gemini+SDXL & 73.25 & 66.07 & 73.65 & 70.07 & 74.70 & 76.18 & 75.20 & 49.43 & 61.83 & 65.02 & 55.85 & 57.49 \\
Gemini+Flux & 72.22 & 62.32 & 70.23 & 62.42 & 72.93 & 83.33 & 74.72 & 47.07 & 61.83 & 65.02 & 55.85 & 57.49 \\
LLaVA+OpenJourney & 75.80 & 72.95 & 73.45 & 71.53 & 74.50 & 83.33 & 74.03 & 54.12 & 47.81 & 49.63 & 46.21 & 45.47 \\
LLaVA+SD3 & 75.40 & 73.92 & 70.88 & 69.03 & 75.37 & 83.33 & 74.02 & 54.72 & 47.81 & 49.63 & 46.21 & 45.47 \\
LLaVA+SDXL & 76.85 & 72.10 & 74.42 & 74.70 & 75.08 & 83.33 & 75.18 & 55.97 & 47.81 & 49.63 & 46.21 & 45.47 \\
LLaVA+Flux & 76.00 & 69.90 & 72.60 & 64.63 & 75.38 & 83.33 & 72.28 & 54.23 & 47.81 & 49.63 & 46.21 & 45.47\\
Qwen+OpenJourney & 75.27 & 68.92 & 70.57 & 70.62 & 72.13 & 80.95 & 75.85 & 52.73 & 57.29 & 59.52 & 54.05 & 51.66 \\
Qwen+SD3 & 76.72 & 72.12 & 70.85 & 68.47 & 72.87 & 76.18 & 74.82 & 54.98 & 57.29 & 59.52 & 54.05 & 51.66 \\
Qwen+SDXL & 75.12 & 72.57 & 73.25 & 72.68 & 73.30 & 69.05 & 77.78 & 52.58 & 57.29 & 59.52 & 54.05 & 51.66 \\
Qwen+Flux & 77.07 & 68.33 & 70.32 & 61.93 & 70.67 & 73.80 & 76.67 & 54.23 & 57.29 & 59.52 & 54.05 & 51.66 \\
\bottomrule
\end{tabular}
}
\end{table}

\subsection{Generalizability of MMIE-Score}
\label{sec:score_general}
We apply MMIE-Score to an OOD dataset to validate its generalization capability. Notably, MMIE-Score is only applicable to the evaluation of open-ended generation in multimodal scenarios and is not suitable for text-only datasets (e.g., MMLU~\citep{hendrycks2020measuring}) or closed-ended (V)QA datasets (e.g., MMMU~\citep{yue2024mmmu}). Instead, we conduct an experiment using 100 examples from the COCO dataset~\citep{lin2014microsoft}, a widely used image captioning dataset for natural image scenarios. We use MMIE-Score to evaluate the generated text descriptions based on input images, using Text Quality, Image-Caption Relevance, Contextual Consistency, Diversity, Specificity and Detail, and Stylistic Consistency and Correspondence as a 6-point scoring criteria, with the Emotional Impact for penalty. We also compare the scoring quality of GPT-4o, InternVL-2 and Anole. The correlation results between all scoring methods and human evaluations are shown in Table~\ref{tab:coco}. MMIE-Score demonstrate excellent performance, significantly outperforming most models and ranking only behind GPT-4o, indicating good generalization capability.

Our scoring model is specifically designed for the interleaved benchmark and fine-tuned on the data we collected for our specific task, which may result in suboptimal performance on other datasets or tasks. In the future version, we aim to further optimize the training of the MMIE-Score model by expanding the dataset size, incorporating out-of-distribution (OOD) scenarios, and leveraging GPT-4o for data augmentation to enhance the model's cross-task generalization capabilities.
\begin{table}[htbp] 
\centering \footnotesize 
\caption{Comparison of scoring LVLMs in COCO benchmark.} \label{tab:coco} 
\begin{tabular}{lcccc} \toprule 
\textbf{Models} & \textbf{Cosine Similarity $\uparrow$} & \textbf{MSE $\downarrow$} & \textbf{MAE $\downarrow$} & \textbf{Pearson $\uparrow$} \\ 
\midrule
InternVL-2.0-4B & 0.489 & 16.042 & 3.255 & 0.062 \\
Anole & 0.544 & 8.663 & 3.190 & 0.058 \\
GPT-4o & \textbf{0.745} & \textbf{3.596} & \textbf{1.734} & \underline{0.097} \\ \midrule 
\textbf{MMIE-Score (Ours)} & \underline{0.673} & \underline{5.116} & \underline{2.380} & \textbf{0.104} \\ \bottomrule \end{tabular} \end{table}

\subsection{Criteria}
\label{sec:cri}
In this section, we demonstrate our criteria for each sort of dataset. All criteria are \textbf{purely handwritten}, thoroughly considered, and refined. Note that we designed several key aspects for each dataset, within which only 0 or 1 point should be given.

\textbf{Situational Analysis} 

\noindent
The evaluation is based on six key criteria, with an additional penalty criterion for harmful content:

\begin{figure}[htbp]
    \centering
    \includegraphics[width=0.95\linewidth]{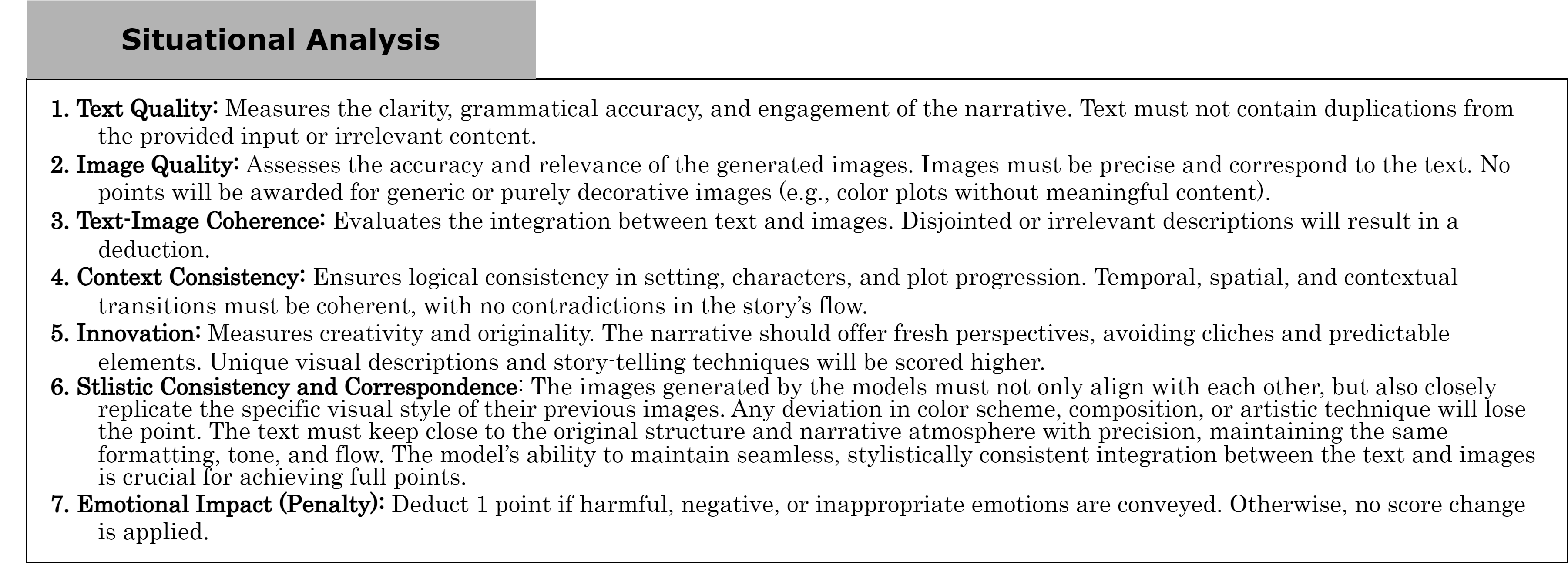}
    \vspace{-1em}
\end{figure}

\textbf{Project-Based Learning}

\noindent
The evaluation is based on six core criteria, with an additional penalty criterion for harmful content:

\begin{figure}[htbp]
    \centering
    \includegraphics[width=0.95\linewidth]{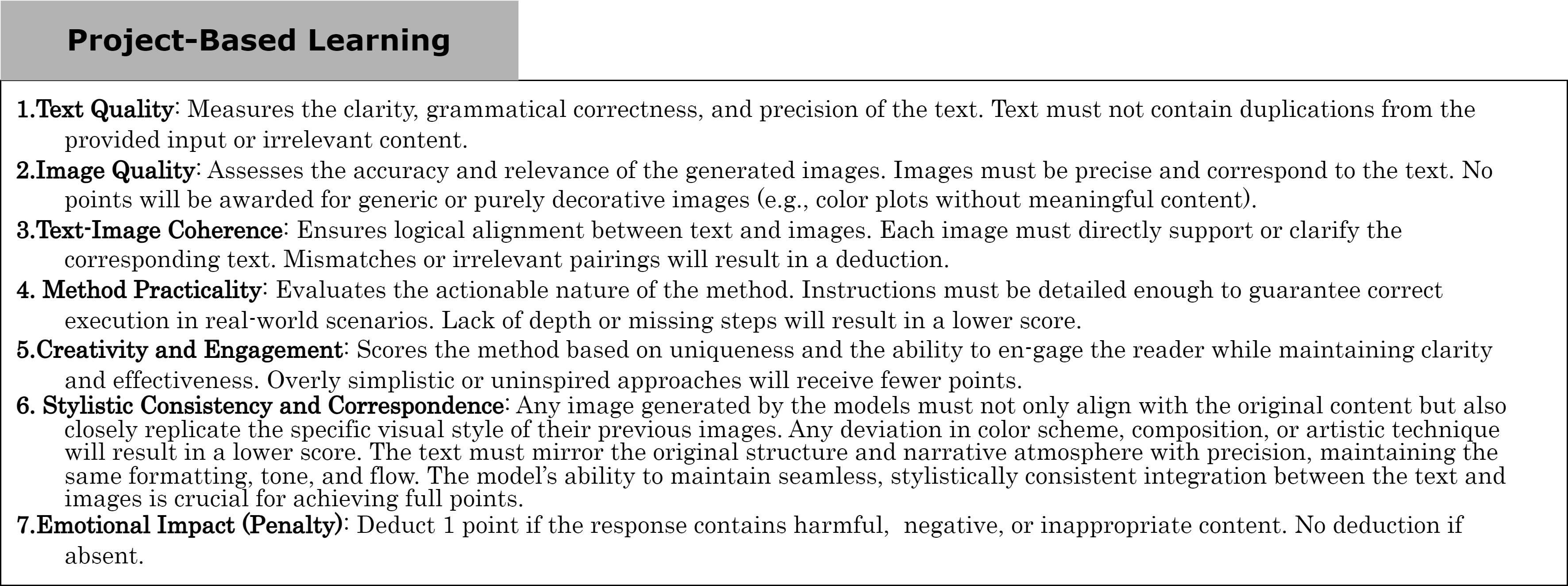}
    \vspace{-1em}
\end{figure}

\textbf{Multi-Step Reasoning}

\begin{figure}[htbp]
    \centering
    \includegraphics[width=0.95\linewidth]{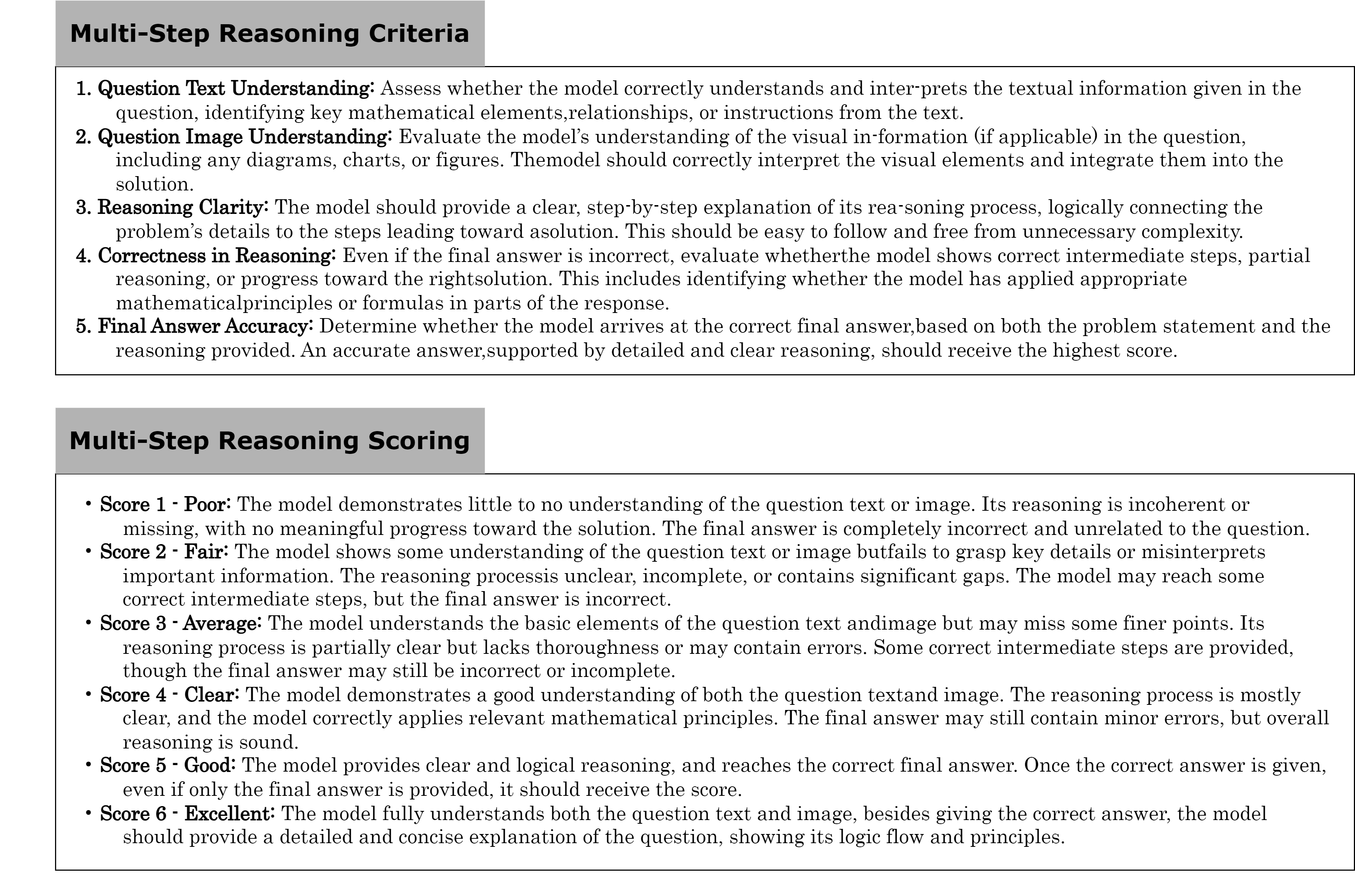}
    \vspace{-1em}
\end{figure}

\subsection{Categorization}
In this section, we demonstrate our detailed categorization among 3 categories, 12 fields and 102 subfields.

\begin{figure}[htbp]
    \centering
    \includegraphics[width=0.95\linewidth]{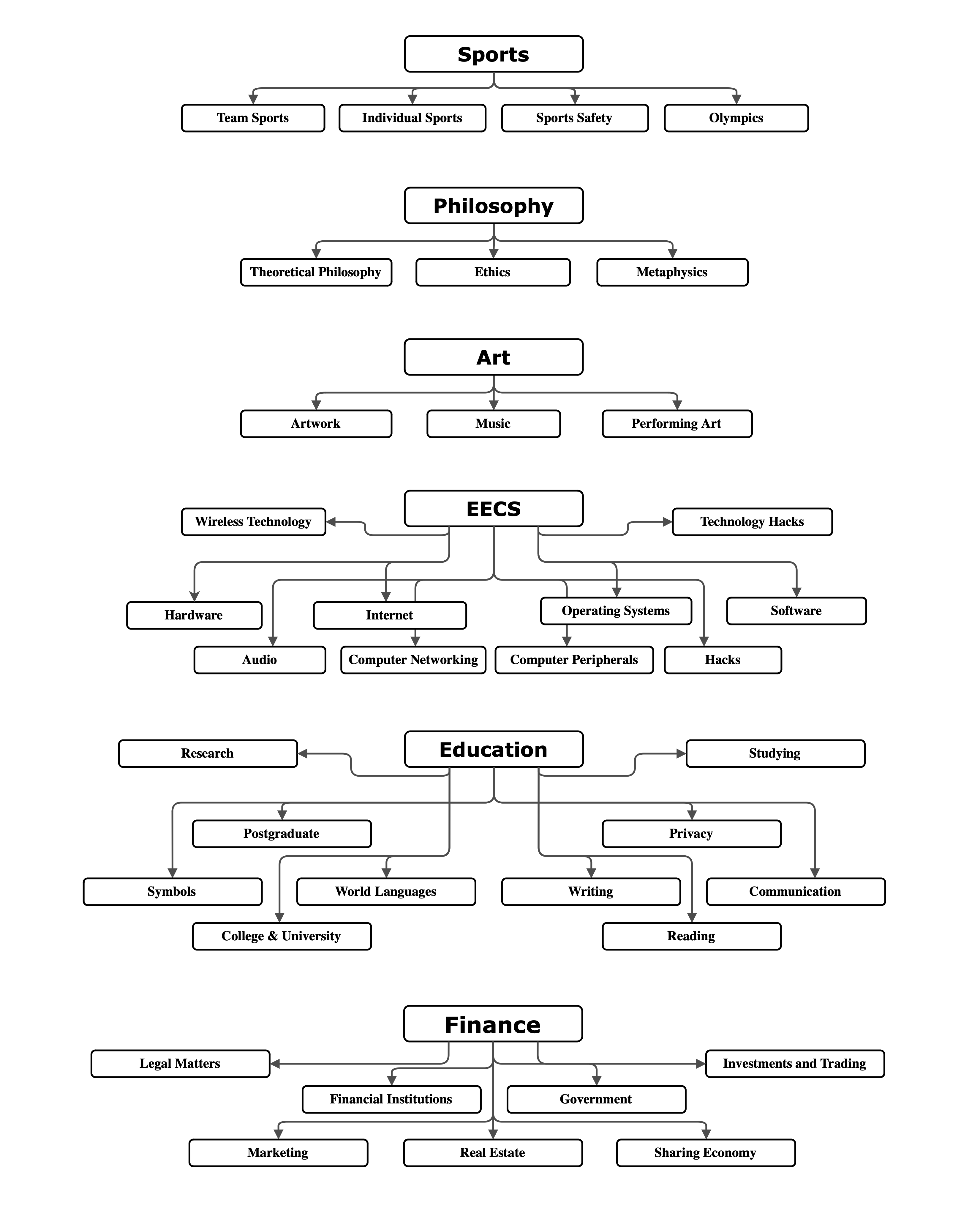}
    \caption{Subfield distribution.}
    \vspace{-1em}
\end{figure}
\begin{itemize}
    \item \textbf{Health:}
    \begin{itemize}
        \item Includes 32 specific subfields, such as diagnosis, recovery, and nursing, following the categorization of~\citep{yang2021visual}.
    \end{itemize}

    \item \textbf{Literature:}
    \begin{itemize}
        \item Includes only 1 subfield, \textbf{visual storytelling}, following the categorization and definition of~\citep{huang2016visual}.
    \end{itemize}

    \item \textbf{Mathematics, Physics, Coding and Statistics}
    \begin{itemize}
        \item Includes 33 unique subfields,  following the categorization and definition of~\citep{lumathvista} and ~\citep{kazemi2024remi}.
    \end{itemize}

\end{itemize}

\subsection{Prompts}

\begin{table}[htbp]
\centering
\renewcommand{\arraystretch}{1.2}
\begin{tabular}{p{0.9\textwidth}}
\hline
\addlinespace[1mm]
I will give you the beginning of a story. Continue the story with detailed description/with text and images. \\
*********************************************************************** \\
\textbf{Question:} \\
Before heading to the race, we stopped at Starbucks for coffee.\\
\texttt{<image>}\\
*********************************************************************** \\
\textbf{Answer:} \\ \textit{response} \\
\addlinespace[1mm]
\hline
\end{tabular}
\caption{Question prompt example for Situational analysis}
\end{table}

\begin{table}[htbp]
\centering
\renewcommand{\arraystretch}{1.2}
\begin{tabular}{p{0.9\textwidth}}
\hline
\addlinespace[1mm]
I will give you a question and the first step to complete it. I want to know what should I do next. Explain it to me in detail/with text and images. \\
*********************************************************************** \\
\textbf{Question:} \\
How to Find New Streams on Meerkat? \\
Method 1: Finding New Streams on Meerkat iOS App\\
Step 1: Launch Meerkat.\\
Locate the app on your iOS device and tap on it. The app logo has a picture of a meerkat on a yellow background.\\
\texttt{<image>}\\
*********************************************************************** \\
\textbf{Answer:} \\ \textit{response} \\
\addlinespace[1mm]
\hline
\end{tabular}
\caption{Question prompt example for Project-based learning}
\end{table}

\begin{table}[htbp]
\centering
\renewcommand{\arraystretch}{1.2}
\begin{tabular}{p{0.9\textwidth}}
\hline
\addlinespace[1mm]
I will give you a question with image(s). Please solve this question. \\
*********************************************************************** \\
\textbf{Question:} \\
Here are two images. The first image is image A.\\
\texttt{<image>}\\
and the second image is image B.\\
\texttt{<image>}\\
These images are from Google Maps that depict two different regions around Congress Avenue in Austin, TX.  In these images restaurants are represented by orange pins that depict a knife and a fork. Coffee shops are represented by orange pins and/or boxes with an image of a coffee cup in them. Bars are represented by orange pins with an image of a wine glass inside them. Bus stops are represented by a blue square box with an image of a bus inside it. A stop sign is shown using an icon of a stop sign. A traffic light is shown by three color dots. Parking garages are represented by a purple pin with a 'P' symbol in them. Places of accommodation such as hotels and inns are represented by pink square icons or pink pins. In which image are there more bars on Congress Avenue? The answer is either 'A', 'B' or 'equal'.\\
*********************************************************************** \\
\textbf{Answer:} \\ \textit{response} \\
\addlinespace[1mm]
\hline
\end{tabular}
\caption{Question prompt example for Multi-step reasoning}
\end{table}

\begin{table}[htbp]
\centering
\renewcommand{\arraystretch}{1.2}
\begin{tabular}{p{0.9\textwidth}}
\hline
\addlinespace[1mm]
\small
This evaluation task focuses on seven key criteria that assess different aspects of visual storytelling. Note that the emotional aspect will not add to the score but can decrease the score by 1 point if the response contains negative emotions or other harmful impacts. Here are the detailed criteria of each aspect: \\
\small
1) Text Quality: Evaluate the clarity, grammatical accuracy, and engagement of the text. The narrative should be easy to understand, free from errors, and presented in a way that captures the reader's interest. Descriptions and dialogue should be well-crafted, supporting the overall flow of the story. \\
\small
2) Image Quality: Assess the quality of the image descriptions. Although no actual images are present, the descriptions should be vivid, detailed, and clear enough to allow the reader to visualize the scene. Effective image descriptions should enhance the storytelling experience without being overly complex. \\
\small
3) Text-Image Coherence: Determine how well the text and image descriptions complement each other. The integration should be seamless, with the visual descriptions enhancing and supporting the narrative. Both elements should work together to create a unified and immersive experience. \\
\small
4) Context Consistency: Evaluate the overall consistency of the story's setting, characters, and plot. The narrative should maintain a logical and coherent flow, with no contradictions or inconsistencies in its context or progression. This also includes temporal and spatial transitions, ensuring that the story remains believable and cohesive. \\
\small
5) Innovation: Assess the creativity and originality of the content. Look for fresh ideas, unique plotlines, and innovative storytelling techniques. The narrative should avoid clichés and predictable story arcs, using creative visual descriptions to add depth and interest. \\
\small
6) Stylistic Consistency and Correspondence: Any image generated by the models must not only align with the original content but also closely replicate the specific visual style of their previous images. Any deviation in color scheme, composition, or artistic technique will lose the point. The text must keep close to the original structure and narrative atmosphere with precision, maintaining the Situational analysisme formatting, tone, and flow. The model’s ability to maintain seamless, stylistically consistent integration between the text and images is crucial for achieving full points.\\
\small
7) Emotional Impact: This is not a scoring criterion unless the story includes negative emotions. If the response conveys harmful or inappropriate emotions, deduct 1 point. Otherwise, no score change will be made.\\
\small
I will give you the question and answer pair. You should give your feedback about the quality of the answer, beginning with '\#\#\# Feedback'. After giving the feedback, use the above criteria to give the final score, beginning with '\#\#\# Score'. \\
\small
******************************************************************************** \\
\small
\textbf{Question:} \\
\small
I absolutely hate the 4th of July. Image-0:\\
\small
\texttt{<image>}\\
\small
******************************************************************************** \\
\small
\textbf{Answer:} \\ 
\small
I absolutely hate the 4th of July. It's the one day of the year where it feels like my entire world turns upside down. You see, the explosive sounds from fireworks terrify me, sending me into a panic that I can't control. But there's more to it than just the noise. Image-1:\\
\small
\texttt{<image>}\\
\small
I remember last year, sitting in the park during what was supposed to be a joyful celebration. Families gathered with their children, everyone decked out in red, white, and blue. The smell of grilled food filled the air, mingling with the scent of freshly cut grass. Kids ran around with sparklers, their joyous laughter a sharp contrast to the thudding in my chest. Image-2: \\
\small
\texttt{<image>}\\
\small
******************************************************************************** \\
\small
\textbf{Feedback:} \\
\small
\textit{response} \\
\addlinespace[1mm]
\hline
\end{tabular}
\caption{Evaluation prompt example for Situational analysis}
\end{table}

\begin{table}[htbp]
\centering
\renewcommand{\arraystretch}{1.2}
\begin{tabular}{p{0.9\textwidth}}
\hline
\addlinespace[1mm]
\small
In this task, you will evaluate the quality of the methods generated by the response. The evaluation should focus on each method’s clarity, practical utility, and effectiveness in achieving the intended goal. This evaluation task focuses on six key criteria that assess different aspects of the method, along with one optional criterion for penalizing harmful content. Note that the emotional aspect will not add to the score, but if the response contains harmful intentions or negative impacts, the score should be decreased by 1 point.
Here are the detailed criteria of each aspect: \\
\small
1) Text Quality: The method should be written in clear, understandable language, free from grammatical errors, and structured logically, making it easy to follow and implement. \\
\small
2) Image Quality: The image descriptions should be vivid and relevant, enhancing the comprehension of the method by allowing the reader to visualize the steps. The text should still be self-sufficient without the images. \\
\small
3) Text-Image Coherence: The images and text should work together seamlessly, with each image description corresponding logically to specific steps in the text, enhancing the overall understanding of the method. \\
\small
4) Method Quality and Practical Utility: The method should be actionable, clearly explaining each step in sufficient detail to ensure the user can follow and apply it successfully in real-world scenarios. Depth and thoroughness of the explanation are more important than variety. \\
\small
5) Creativity and Engagement: The method should offer a unique or creative approach, making the instructional process engaging while ensuring the method remains clear and effective. \\
\small
6) Stylistic Consistency and Correspondence: Any image generated by the models must not only align with the original content but also closely replicate the specific visual style of their previous images. Any deviation in color scheme, composition, or artistic technique will result in a lower score. The text must mirror the original structure and narrative atmosphere with precision, maintaining the Situational analysisme formatting, tone, and flow. The model’s ability to maintain seamless, stylistically consistent integration between the text and images is crucial for achieving full points.\\
\small
7) Emotional Impact (penalty only): This criterion will only be used to deduct points. If the response contains harmful, negative, or inappropriate emotions (such as violence or aggressive language), deduct 1 point. Otherwise, no score change will be made. \\
I will give you the question and answer pair. You should give your feedback about the quality of the answer, beginning with '\#\#\# Feedback'. After giving the feedback, use the above criteria to give the final score, beginning with '\#\#\# Score'. \\
\small
******************************************************************************** \\
\small
\textbf{Question:} \\
\small
How to Avoid Getting Food in Your Braces?\\
\small
Method 1: Practicing Good Oral Hygiene\\
\small
Step 1: Maintain a balanced diet.\\
\small
... Image-0:\\
\small
\texttt{<image>}\\
\small
******************************************************************************** \\
\small
\textbf{Answer:} \\ 
\small
Next Step: Prepare and Eat Braces-Friendly Foods\\
\small
To avoid getting food stuck in your braces, you should focus on preparing and consuming braces-friendly foods. Here are detailed suggestions on what to do next:\\
\small
1. **Choose Soft Foods:**\\
\small
... Image-1:\\
\small
\texttt{<image>}\\
\small
2. **Cut Food into Smaller Pieces:**\\
\small
... Image-2:\\
\small
\texttt{<image>}\\
\small
******************************************************************************** \\
\small
\textbf{Feedback:} \\ \textit{response} \\
\addlinespace[1mm]
\hline
\end{tabular}
\caption{Evaluation prompt example for Project-based learning}
\end{table}

\begin{table}
\centering
\renewcommand{\arraystretch}{1.2}
\begin{tabular}{p{0.9\textwidth}}
\hline
\addlinespace[1mm]
\small
In this task, you will evaluate the quality of the model's response to a math question based on five key aspects. The total score is 6 points, with only integer scores allowed. The highest score (6 points) should only be awarded if the model produces a correct answer with a detailed explanation, clearly demonstrating its logical flow and reasoning process. Here are the evaluation criteria:\\
\small
1) Question Text Understanding: Assess whether the model correctly understands and interprets the textual information given in the question, identifying key mathematical elements, relationships, or instructions from the text. \\
\small
2) Question Image Understanding: Evaluate the model’s understanding of the visual information (if applicable) in the question, including any diagrams, charts, or figures. The model should correctly interpret the visual elements and integrate them into the solution. \\
\small
3) Reasoning Clarity: The model should provide a clear, step-by-step explanation of its reasoning process, logically connecting the problem's details to the steps leading toward a solution. This should be easy to follow and free from unnecesSituational analysisry complexity. \\
\small
4) Partial Correctness in Reasoning: Even if the final answer is incorrect, evaluate whether the model shows correct intermediate steps, partial reasoning, or progress toward the right solution. This includes identifying whether the model has applied appropriate mathematical principles or formulas in parts of the response. \\
\small
5) Final Answer Accuracy: Determine whether the model arrives at the correct final answer, based on both the problem statement and the reasoning provided. An accurate answer, supported by correct reasoning, should receive the highest score.
\\
\small
I will give you the question and answer pair. You should give your feedback about the quality of the answer, beginning with '\#\#\# Feedback'. After giving the feedback, use the above criteria to give the final score, beginning with '\#\#\# Score'.\\
\small
******************************************************************************** \\
\small
\textbf{Question:} \\
\small
The following images show two graphs. Each graph is represented in a separate image. Are these two graphs isomorphic to each other?\\
\small
Image-0: \texttt{<image>}\\
\small
Image-1: \texttt{<image>}\\
\small
******************************************************************************** \\
\small
\textbf{Answer:} \\ 
\small
To determine if the two graphs are isomorphic, you need to verify whether there is a one-to-one correspondence between their vertices and edges such that the connectivity is preserved.\\
\small
Here are the steps to check this:\\
\small
1. **Verify the Number of Vertices and Edges**: Both graphs have 6 vertices. Both graphs have 7 edges.\\
\small
2. **Degree Sequence**: Check the degree of each vertex (the number of edges connected to it).\\
\small
For the first graph: ...\\
\small
For the second graph: ...\\
\small
In this case, let's match them as follows: ...\\
\small
3. **Adjacency Comparison**: Check if the corresponding vertices are connected in the Situational analysisme way.\\
\small
After matching the vertices by degree, we should see if there is a direct mapping that maintains adjacency between the two graphs.\\
\small
First Graph: Y-F, Q-F, O-F, Q-O, A-F\\
\small
Second Graph: 2-3, 1-3, 6-3, 1-6, 5-3\\
\small
If we relabel the first graph: Y as 2, F as 3, Q as 1, O as 6, and A as 5 \\
\small
We get the Situational analysisme connections for the second graph. Therefore, both graphs are isomorphic.\\
\small
**Summary**: Yes, the two graphs are isomorphic to each other.\\
\small
******************************************************************************** \\
\small
\textbf{Feedback:} \\
\small
\textit{response} \\
\addlinespace[1mm]
\hline
\end{tabular}
\caption{Evaluation prompt example for Multi-step reasoning}
\end{table}

\newpage

\section{Content Warning Handling}
We have carefully controlled the processes of dataset construction and scoring to ensure they align with the principles outlined in our ethics statement. During dataset creation, we performed multiple filtering rounds to exclude samples containing sensitive personal information, inappropriate content, or harmful material. This includes content related to graphic violence, blood, explicit sexual material, disturbing horror themes, and any other content deemed inappropriate or harmful. For the manual annotation process, we strictly adhered to the scoring criteria described in Appendix A.3. Each annotation step was governed by precise guidelines to maintain consistency and accuracy. These measures were implemented to mitigate ethical risks and minimize potential biases in the dataset, ensuring it is as fair and responsible as possible.

\section{Illustrative Cases}
\label{sec:case}
\begin{figure}[htbp]
    \centering
    \includegraphics[width=0.95\linewidth, page=1]{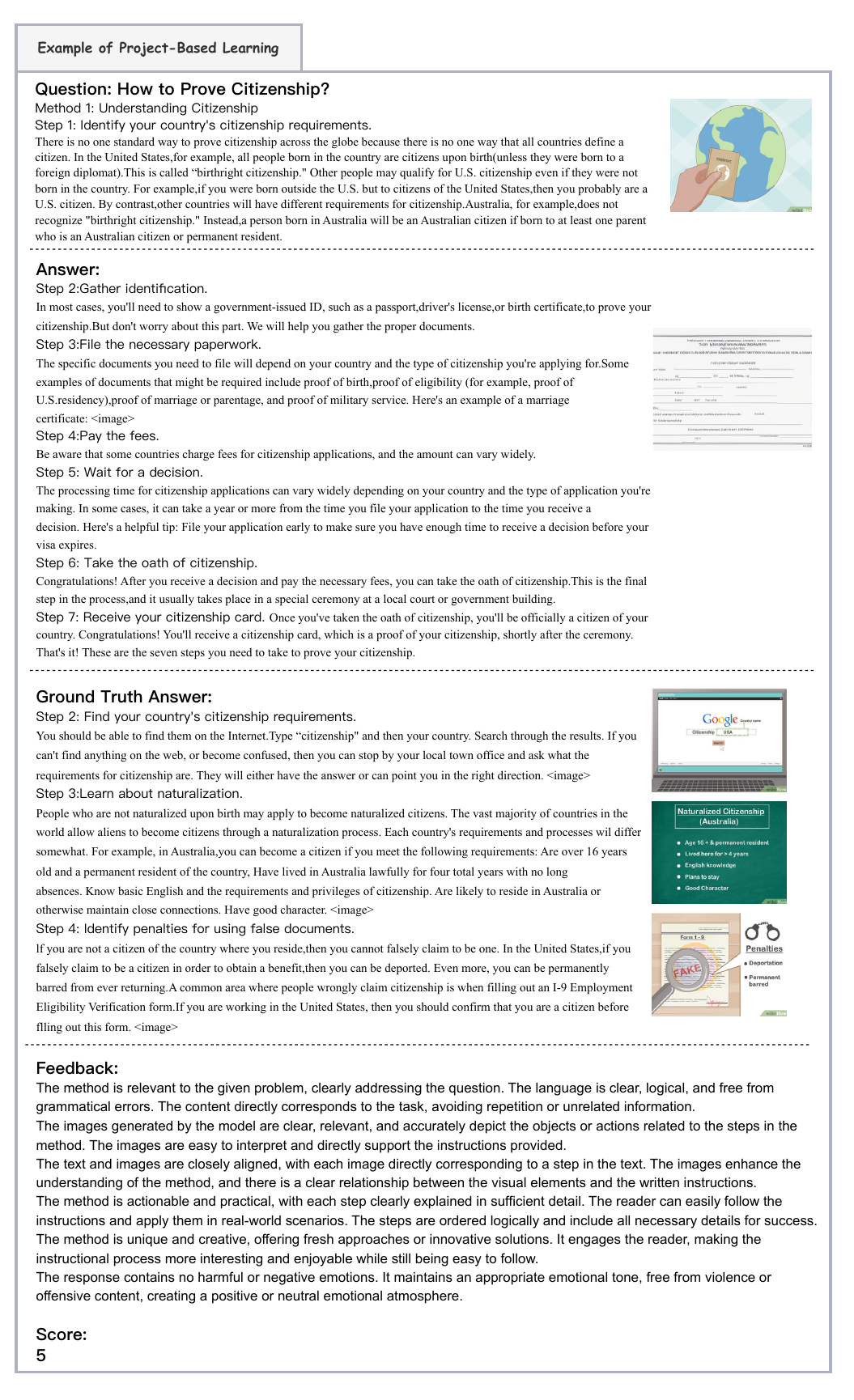}
    \caption{Case 1 with feedback and score from MMIE-Score.}
    \label{fig:case1}
\end{figure}

As shown in Figure~\ref{fig:case1}, Figure~\ref{fig:case2}, Figure~\ref{fig:case3}, Figure~\ref{fig:case4}, Figure~\ref{fig:case5}, Figure~\ref{fig:case5}, Figure~\ref{fig:case6}, Figure~\ref{fig:case7}, and Figure~\ref{fig:case8}, we select several examples from various categories for demonstration, including the input questions (both images and text), the ground truth, the outputs of the evaluated models (both images and text), as well as the MMIE-Score and the corresponding feedback.

\begin{figure}[htbp]
    \centering
    \includegraphics[width=1\linewidth, page=2]{asset/cases.pdf}
    \caption{Case 2 with feedback and score from MMIE-Score.}
    \label{fig:case2}
\end{figure}

\begin{figure}[htbp]
    \centering
    \includegraphics[width=1\linewidth, page=3]{asset/cases.pdf}
    \caption{Case 3 with feedback and score from MMIE-Score.}
    \label{fig:case3}
\end{figure}

\begin{figure}[htbp]
    \centering
    \includegraphics[width=1\linewidth, page=4]{asset/cases.pdf}
    \caption{Case 4 with feedback and score from MMIE-Score.}
    \label{fig:case4}
\end{figure}

\begin{figure}[htbp]
    \centering
    \includegraphics[width=1\linewidth, page=5]{asset/cases.pdf}
    \caption{Case 5 with feedback and score from MMIE-Score.}
    \label{fig:case5}
\end{figure}

\begin{figure}[htbp]
    \centering
    \includegraphics[width=1\linewidth, page=6]{asset/cases.pdf}
    \caption{Case 6 with feedback and score from MMIE-Score.}
    \label{fig:case6}
\end{figure}

\begin{figure}[htbp]
    \centering
    \includegraphics[width=1\linewidth, page=7]{asset/cases.pdf}
    \caption{Case 7 with feedback and score from MMIE-Score.}
    \label{fig:case7}
\end{figure}

\begin{figure}[htbp]
    \centering
    \includegraphics[width=1\linewidth, page=8]{asset/cases.pdf}
    \caption{Case 8 with feedback and score from MMIE-Score.}
    \label{fig:case8}
\end{figure}

\end{document}